\documentclass[preprint,12pt]{elsarticle}

\usepackage{amssymb}
\usepackage{url}
\usepackage{subfigure}
\usepackage{amsmath}
\usepackage{threeparttable}
\usepackage{algorithm}
\usepackage{algorithmic}
\usepackage{epstopdf}
\usepackage{booktabs}
\usepackage{color}
\usepackage{multirow}
\usepackage{caption}
\usepackage{mathrsfs}
\usepackage{appendix}
\newtheorem{definition}{Definition}
\newtheorem{theorem}{Theorem}

\newtheorem{prop}{Proposition}
\newtheorem{lemma}{Lemma}[section]
\biboptions{numbers,sort&compress}
\newtheorem{example}{Example}[section]
\newtheorem{remark}{Remark}[section]
\newcommand{\be}{\begin{equation}}
\newcommand{\ee}{\end{equation}}
\newcommand{\ba}{\begin{eqnarray}}
\newcommand{\ea}{\end{eqnarray}}
\newcommand{\bas}{\begin{eqnarray*}}
	\newcommand{\eas}{\end{eqnarray*}}

\def\libsvc{\texttt{LIBSVC}}
\def\lssvc{\texttt{LSSVC}}
\def\l2svc{\texttt{L2SVC}}
\def\ramp{\texttt{RAMP}}
\def\rsvc{\texttt{RSVC}}
\def\rshsvc{\texttt{RSHSVC}}

\def\mACC{\texttt{mACC}}
\def\mCPU{\texttt{mCPU}}
\def\mNSV{\texttt{mNSV}}
\usepackage{lineno}

\renewcommand\tabcolsep{0.5pt}
\journal{}
\begin{document}
\begin{frontmatter}
%% Title, authors and addresses

%% use the tnoteref command within \title for footnotes;
%% use the tnotetext command for theassociated footnote;
%% use the fnref command within \author or \address for footnotes;
%% use the fntext command for theassociated footnote;
%% use the corref command within \author for corresponding author footnotes;
%% use the cortext command for theassociated footnote;
%% use the ead command for the email address,
%% and the form \ead[url] for the home page:
%% \title{Title\tnoteref{label1}}
%% \tnotetext[label1]{}
%% \author{Name\corref{cor1}\fnref{label2}}
%% \ead{email address}
%% \ead[url]{home page}
%% \fntext[label2]{}
%% \cortext[cor1]{}
%% \address{Address\fnref{label3}}
%% \fntext[label3]{}
\title{Nonlinear Kernel Support Vector Machine with  0-1 Soft Margin Loss}
%\tnotetext[label1]{This work is supported by the Natural Science Foundation of Hainan Province (No.118QN181), the National Natural Science Foundation of China (No.61703370, No.11871183, No.61866010, No.11501310 and No.61603338), and the Natural Science Foundation of Zhejiang Province (No.LQ17F030003, No.LY18G010018 and No.LY16A010020).}

%% use optional labels to link authors explicitly to addresses:
%% \author[label1,label2]{}
%% \address[label1]{}
%% \address[label2]{}

%\cortext[cor1]{Corresponding
%author.}

\author[1]{Ju Liu}
\address[1]{Management School, Hainan University, Haikou, 570228, P.R.China}

\author[1]{Ling-Wei Huang}

\author[1]{Yuan-Hai Shao\corref{cor1}}
\cortext[cor1]{Corresponding author.}\ead{shaoyuanhai21@163.com}

\author[3]{Wei-Jie Chen}
\address[3]{Zhijiang College, Zhejiang University of Technology, Hangzhou,
310024, P.R.China}

\author[1]{Chun-Na Li}

%\address{}
\begin{abstract}
Recent advance in linear support vector machine with the 0-1 soft margin loss ($L_{0/1}$-SVM) shows  the probability of solving the 0-1 loss problem directly. However, its theoretical and algorithmic requirements restrict us from directly extending the linear solving framework to its nonlinear kernel form. The lack of an explicit expression of the Lagrangian dual function of $L_{0/1}$-SVM is a major shortcoming among them. By applying the nonparametric representation theorem, we propose a nonlinear model for support vector machine with  0-1 soft margin loss, called $L_{0/1}$-KSVM, which skillfully incorporates the kernel technique,  and more importantly, follows the success in systematically solving its linear problem. The optimal condition is theoretically explored and a working set selection alternating direction method of multipliers (ADMM)  algorithm is introduced to obtain its numerical solution. Furthermore, we first introduce a closed-form definition to the support vector (SV) of $L_{0/1}$-KSVM. Theoretically,
we prove that all SVs of $L_{0/1}$-KSVM are  only located on the parallel decision surfaces. The experimental results show that $L_{0/1}$-KSVM has much fewer SVs compared to its linear counterpart and the other six nonlinear benchmark SVM classifiers, while maintaining good prediction accuracy.

\end{abstract}

\end{frontmatter}

%\linenumbers
%${\textbf{X}} \times \R = \{(\textbf{x}_1,y_1), (\textbf{x}_2,y_2),~\ldots~,(\textbf{x}_m,y_m)\} \in\R^{n}\times \{\pm{1}\}$
\section{Introduction}
Vapnik and Cortes first proposed the support vector machine (SVM) for binary classification in 1995 [1], then it became a very popular modelling and prediction tool for many small and moderate scale machine learning, statistic and pattern recognition problems [2]. The SVM classifier aims to build a linear or nonlinear maximum-margin decision boundary to separate two classes of data points.
Usually, the nonlinear SVM model is  preferred  in most practical scenarios [3]. Given a data set
${\mathcal{X}} \times {\mathbb{Y}}=\{(\textbf{x}_1,y_1), (\textbf{x}_2,y_2),~\ldots~,(\textbf{x}_m,y_m)\} \subseteq\mathbb{R}^n\times \{{+1},{-1}\}$, here $y_{i}\in\mathbb{Y}=\{{+1},{-1}\} $$(i \in {\mathbb{N}}_{m}, {\mathbb{N}}_{m}:=1,~\cdots~,m)$ is the label value corresponding to input $\textbf{x}_{i}\in {\mathbb{R}}^n(i \in {\mathbb{N}}_{m})$, the formulation of the nonlinear SVM  problem is

\begin{eqnarray} \label{reformulate000}
&\underset{ \textbf{w} \in {\mathbb{R}}^n,b \in \mathbb{R}}{\min} & \frac{1}{2} \Vert \textbf{w}  \Vert^2+ C\sum_{i=1}^{m}\ell(1-y_{i}f(\textbf{x}_{i})) \nonumber
\end{eqnarray}
where $C>0$ is a penalty parameter and the decision surface ${f}(\cdot)$ is defined by  ${f}({\textbf{x}})=\left \langle {\textbf{w}},\Phi(\textbf{x}) \right \rangle+b=0$,  $\Phi(\cdot)$ is an appropriate feature map that expects new higher-dimensional points $\Phi(\textbf{x})(\textbf{x} \in {\mathcal{X}})$ linearly separable in the feature space.
The loss function $\ell(\cdot)$ measures the penalty  on the misclassified data points $\{\textbf{x}_{i} \bigg|1-y_{i}f(\textbf{x}_{i})>0, i \in {\mathbb{N}}_{m}\}$. The 0-1 loss function $\ell_{0/1}(\cdot)$ is  a natural option of $\ell(\cdot)$ since it captures the discrete nature of binary classification, i.e., it is designed to minimize the number of misclassified instances
\begin{eqnarray*}
\ell_{0/1}(u_i)=
\begin{cases}
1,& u_i>0,\\
0,& u_i\leq0,
\end{cases}
\end{eqnarray*}
here $u_i=1-y_{i}f(\textbf{x}_{i}), i \in {\mathbb{N}}_{m}$.
However, its non-convexity and discontinuity  make  the problems with 0-1 loss   NP-hard to optimize [4-7]. For decades, compromising with the challenge of directly solving the original SVM with 0-1 loss, many researchers transfer to operate the following two alternative strategies instead: replacing the  0-1 loss function with surrogate loss functions,  or executing approximate algorithms based on the binary-valued property of  0-1 loss function.

\begin{figure}[htpb]
\begin{center}{
{\resizebox*{11cm}{!}
{\includegraphics{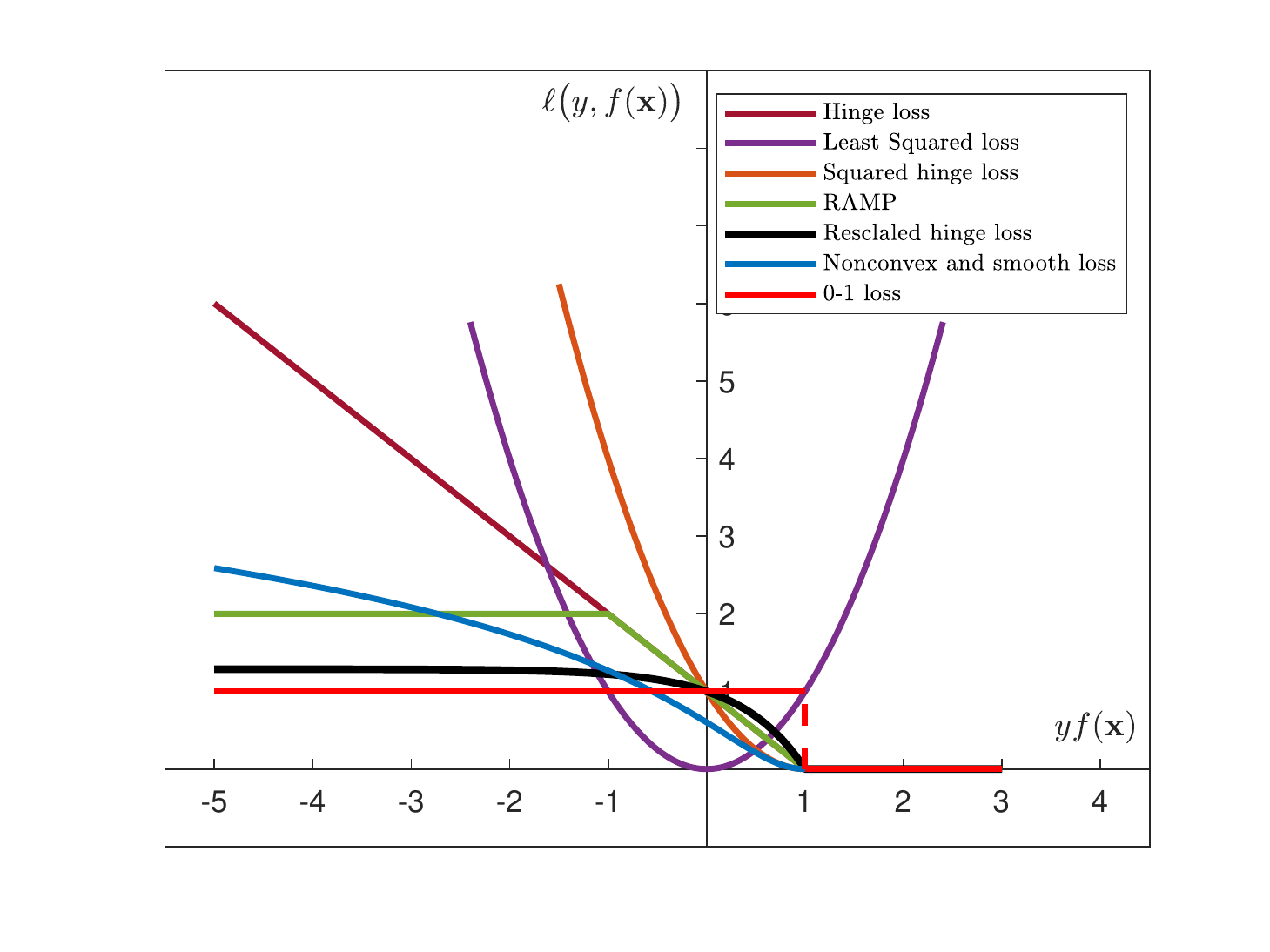}}}\hspace{3pt}
\caption{The 0-1 loss function and other common loss functions.}
\label{fig:rho-0}}
\end{center}
\end{figure}

In the SVM community,  a tremendous amount of work has been devoted to the first strategy, in particular to SVMs with convex surrogates because  of their ease-of-use in practical applications and their convenience in theoretical analyses. However, there is  no solid evidence on  how well these convex surrogates approximate the  0-1 loss itself, even if they have asymptotic relevance [8].
Among the most widely researched surrogates are  hinge loss [1], least squares loss [9], squared hinge loss [10],
pinball loss [11] and  log loss [12]. Compared to convex surrogates,  ramp loss [13,14], the rescaled hinge loss [15], the non-convex and smooth loss  [16] (See Figure \ref{fig:rho-0}), etc, these non-convex surrogates still attract much attention in certain scenes requiring high insensitivity to outliers. At the same time, these non-convex SVMs show relatively better performance in classification accuracy and the SVs decline. As for the majority of nonlinear SVMs with convex surrogates, their optimization problems can be systematically researched by the dual kernel-based technique combined with KKT condition. Unfortunately, this method cannot guarantee the same success for many non-convex surrogates. With respect to the  0-1 loss, the ideal candidate for SVM loss, we find it is challenging to implement the  Lagrangian dual approach directly. The reason is that  the core function, namely, the Lagrangian dual function of the 0-1 loss SVM can not be explicitly expressed due to the  non-smoothness and non-convexity of 0-1 loss function.
 Finally, the failure of this popular technique, also the lack of  other effective ways to directly solve this NP-hard problem
 makes the task of exactly solving the 0-1 loss SVM a decades-long  challenge in the SVM community.

As for the second strategy, some approximate algorithms  used sigmoid function [17] or a sigmoidal
transfer function [18] as the simulation of 0-1 loss for their similar numerical characteristics. Moreover, Orsenigo and Vercellis [19] proposed the algorithm of multivariate classification trees based on minimum features discrete support vector machines and an approximate technique was performed to solve the mixed integer problem (MIP) model at each node of trees in their method.  A point must be acknowledged that these approximate algorithms mainly focus on  techniques other than systematic solving framework including sound theoretical analysis, such as proposing an optimal condition, and the  solving algorithm more closely adhering to the SVM problems.

  Most lately, Wang et al. [20] have made a great step in their systematic work aimed at directly solving the linear $L_{0/1}$-SVM, an abbreviation  for the linear SVM with 0-1 loss. To the best  of our  knowledge, they first  built the optimal condition for both the global and local solutions of the linear $L_{0/1}$-SVM by virtue of proximal operator of 0-1 loss function, and then  an efficient algorithm $L_{0/1}$-ADMM was  invented to find the approximate solution based on the optimal condition. In view of the fundamental or theoretical achievements on directly solving the  SVM with 0-1 loss are not yet fully developed,
 the relatively systematic work  on linear $L_{0/1}$-SVM  in [20] is a meaningful attempt on this topic. Therefore, we expect that their inspiring view  can lead us to build a similar systematic framework for nonlinear $L_{0/1}$-SVM.

 A powerful and popular method for dealing with the nonlinear SVMs is the kernel method [9][21].  In general, most nonlinear SVM problems need to incorporate the kernel technique into their Lagrangian dual functions. As mentioned earlier,  the explicit formulation of the Lagrangian dual function of $L_{0/1}$-SVM is unaccessible because the 0-1 loss function has the negative property of convexity and continuity. The dilemma forces researchers to explore other ways that can still effectively use kernel trick but without involving the surrounding of the Lagrangian dual problem. One way is  to approximate the kernel function by proper feature map. The technique for speeding up kernel methods on large-scale problems, called the class of random features [22,23],  is of this type. Specifically, it approximately decomposes the kernel function by using a explicit feature map  and then transforms the nonlinear SVM problems into the linear case.

 In this paper, based on the nonparametric representation theorem [24], a  nonlinear $L_{0/1}$-SVM model, called $L_{0/1}$-KSVM is presented. The new $L_{0/1}$-KSVM model skillfully incorporates the kernel technique  without using the Lagrangian dual function. More importantly,
 the proposed model also successfully extend the systematic solving framework of linear model in [20], in other words,
  the corresponding optimal condition and algorithm are similarly explored, as the linear framework was completed in [20]. The main contributions can be summarized as follows.\\
(i) A new formulation  of the nonlinear kernel $L_{0/1}$-SVM, called  $L_{0/1}$-KSVM, is presented. The new proposed $L_{0/1}$-KSVM simultaneously  incorporates the kernel technique and adopts the advanced and systematic solving framework of $L_{0/1}$-SVM. Moreover, we  prove that the $L_{0/1}$-KSVM model    degenerates into a slightly modified edition of the original linear $L_{0/1}$-SVM  by adding an extra regularization $\frac{1}{2} b^2$ into its objective function.\\
(ii) A precise definition of the support vector and the optimality condition of the $L_{0/1}$-KSVM are given by the proximal constraints of P-stationary point and  an important conclusion about SVs is also explored:  SVs are only located on the parallel decision surfaces ${f(\cdot)}= \pm1$, where ${f(\cdot)}=0$ is the final decision surface. This geometric character relating to SVs  echoes the fact that SVMs with 0-1 loss has much smaller SVs  than other SVM peers with traditional convex losses. \\
(iii) The experimental part, respectively using $L_{0/1}$-SVM and the other six classical nonlinear SVMs for performance comparison, has validated the strong sparsity and reasonable robustness of our proposed $L_{0/1}$-KSVM. The result of much fewer SVs of the $L_{0/1}$-KSVM is in stark contrast to all other experimental counterparts. Meanwhile, we have achieved a decent result in terms of prediction accuracy compared to  the other six classical nonlinear SVMs.

The rest of the paper is organized as follows. Section \ref{pk21new} gives a broad outline for the $L_{0/1}$-SVM and then presents two foundations for modelling the nonlinear $L_{0/1}$-KSVM, the kernel method and the representation theorem. The whole framework of  $L_{0/1}$-KSVM, which is also the topic of the current paper, is explicitly studied in section \ref{fnlsvm0}. Finally in section
\ref{fnlsvm088}, we will prove the advantageous aspects of $L_{0/1}$-KSVM  on strong sparsity and robustness through experiments compared to the $L_{0/1}$-SVM and the other six classical nonlinear SVMs.

\section{ Preliminary knowledge}\label{pk21new}
First, we give a brief overview on linear $L_{0/1}$-SVM; then the following  part focuses on presenting the knowledge about
the kernel method and the representation theorem,  which are crucial for building our final model $L_{0/1}$-KSVM.
For convenience, some notations are given in Table \ref{TableUCIInfo}.
\begin{table*}[htbp]
\begin{center}
\renewcommand\tabcolsep{1.0pt}
\caption{List of notations.}
\resizebox{3.5in}{!}
{
\begin{tabular}{lccccclccccc}
\toprule
{\multirow{1}{*}{Notation}}& \vline &Description\\
\midrule
$\textbf{X}$&\vline&$[\textbf{x}_1 \ \textbf{x}_2 \ldots \ \textbf{x}_m]^\top\in\mathbb{R}^{m\times n}$ \\
$A$&\vline&$[y_1\textbf{x}_{1}~y_2\textbf{x}_{2}~\ldots~y_m\textbf{x}_{m}]^\top\in\mathbb{R}^{m\times n}$ \\
$\textbf{y}$&\vline&$(y_{1},y_2,\ldots,y_{m})^{\top}\in \mathbb{R}^{m}$ \\
$\widetilde{A}$&\vline&$[A,\textbf{y}]\in \mathbb{R}^{m\times (n+1)}$ \\
$ \textbf{V}$&\vline&$\text{Diag}(\textbf{y})=\text{Diag}(y_{1},y_2,\ldots,y_{m})\in {\mathbb{R}}^{m\times m}$ \\
$ {\textbf{e}}$&\vline&$(1,1,\ldots,1)^{\top}\in {\mathbb{R}}^{m}$ \\
$ {\widetilde{{A}}}^+$&\vline&$({\widetilde{{A}}}^\top {\widetilde{{A}}})^{-1}{\widetilde{{A}}}^\top \in {\mathbb{R}}^{(n+1)\times m}$ \\
$ H$&\vline&$\begin{bmatrix}
I_{n\times n} & {\bf0}  \nonumber\\
{\bf0} & 0 \nonumber\\
\end{bmatrix}{\widetilde{{A}}}^+\in {\mathbb{R}}^{(n+1)\times m}$, here ${\bf0} \in {\mathbb{R}}^n$ \\
$ t_+$&\vline&$\text{max}\{t, 0\}, t \in {\mathbb{R}}$ \\
$ {\textbf{u}}$&\vline&$(u_{1},u_2,\ldots,u_{m})^{\top}\in {\mathbb{R}}^{m}$ \\
$ {\textbf{u}}_+$&\vline&$((u_1)_+, \ldots,(u_m)_+)^\top \in {\mathbb{R}}^{m}$ \\
$ \|\textbf{u}\|_0$&\vline&The number of non-zero elements in $\textbf{u}$ \\
\bottomrule
\end{tabular}
}
\label{TableUCIInfo}
\end{center}
\end{table*}

\subsection{Linear $ L_{0/1}$-SVM}\label{pk21new000}
We now give a simple overview related to the  $L_{0/1}$-SVM   by Wang and his partners in [20]. First, they rewrote the SVM with 0-1 loss  into the following matrix form
\begin{equation}\label{reformulate}
\begin{split}
\underset{ \textbf{w} \in {\mathbb{R}}^n,b \in \mathbb{R},{\textbf{u}} \in {\mathbb{R}}^m}{\min} & ~~\frac{1}{2} \Vert \textbf{w}  \Vert^2+C\|{\textbf{u}}_{+}\|_{0}\\
\mbox{s.t.\ }& ~~{\textbf{u}}+A  \textbf{w} +b\textbf{y}={\textbf{e}},
\end{split}
\end{equation}
and the hyperplane or decision function is of the form  ${f}({\textbf{x}})=\left \langle {\textbf{w}},{\textbf{x}} \right \rangle$+$b$.     Note that $\textbf{u}$  is obtained here under the condition that $\Phi$ is an identity map.

The optimal condition of \eqref{reformulate} was presented in   Theorem \ref{gol-p0}, which was introduced by an important concept, the P-stationary point. We will review them respectively in the following part:
\begin{definition}

[20][P-stationary point of \eqref{reformulate}]\label{pstationarypoint} For a given $C>0$, we say   $({ \textbf{w}^* }; b^*;  \textbf{u}^* )$ is a proximal stationary (P-stationary) point of \eqref{reformulate} if  there are a Lagrangian multiplier vector $ {\boldsymbol{\lambda}^*}\in \mathbb{R}^m$ and a constant $\gamma>0$ such that

\be\label{aaalfi0}
\left\{
\begin{array}{rll}   \textbf{w}^* + A^{\top}{\boldsymbol{\lambda}^*}&={\bf 0},\\
\langle \textbf{y}, {\boldsymbol{\lambda}^*} \rangle&= {\bf 0},\\
 {\textbf{u}}^*+A   \textbf{w}^*+b^*\textbf{y}&={\textbf{e}},\\
\text{Prox}_{\gamma C\|({\cdot})_{+}\|_{0}}(\textbf{u}^*-\gamma{\boldsymbol{\lambda}^*})&=\textbf{u}^*,
\end{array}
\right.
\ee
where
\begin{equation}
\label{exp-proximal1LL01}
[\text{Prox}_{\gamma C\|({\cdot})_{+}\|_{0}}({ \textbf{z}^*})]_i=
\begin{cases}
0,& 0< z^*_i\leq\sqrt{2\gamma C},\\
z^*_i,& z^*_i>\sqrt{2\gamma C}~ \text{or}~z^*_i\leq0 .
\end{cases}
\end{equation}
Here, $\textbf{z}^*:=\textbf{u}^*-\gamma{\boldsymbol{\lambda}^*}$.\end{definition}
\begin{theorem}\label{gol-p0}[20]  For problem \eqref{reformulate}, the following relations hold.

(i)~ Assume $\widetilde{{A}}$ has a  full  column rank. For a  given $C>0$, if $(  \textbf{w}^*; b^*; {\textbf{u}^*})$ is a global minimizer of \eqref{reformulate}, then it is a P-stationary point of \eqref{reformulate} with $0<\gamma< 1/\lambda_H$, where $\lambda_H$ represents the maximum eigenvalue of $H^\top H.$\label{gol-prox1}

(ii)~For a  given $C>0$, if $(  \textbf{w}^*; b^*; {\textbf{u}^*})$ is a P-stationary point of \eqref{reformulate} with $\gamma>0$, then it is a local minimizer of \eqref{reformulate}.\label{gol-prox2}
\end{theorem}

The theorem states that the P-stationary point must be a local minimizer under a certain condition, and  [20] used the P-stationary point as a termination rule for the iterative points generated by {$L_{0/1}$-ADMM}, the algorithm for numerically solving \eqref{reformulate}.
 In addition, they introduced  a new expression to the SV based on  P-stationary point and further verified that SVs lie only on the parallel decisive hyperplanes ${f}({\textbf{x}})=\left \langle {\textbf{w}},{\textbf{x}} \right \rangle$+$b= \pm1$, which was expected to guarantee the high efficiency because of its strong sparsity and relative robustness due to its insensitivity to outliers.

In [20], the $L_{0/1}$-SVM in theory and experiment has shown the great advantage of  rather a
shorter computation time and many fewer SVs  compared
to other leading linear SVM classifiers. However, we point out some obstacles when we extend the whole efficient modelling framework from the linear form to the nonlinear form.\\
     (i). Theoretically,  the number of the representation of the nonlinear or kernel basis is always larger than the number of training points. Therefore, the requirement that the full  column rank of $\widetilde{{A}}$ in Theorem \ref{gol-p0} in [20] cannot be satisfied if we normally introduce nonlinear mapping or kernel trick to handle the nonlinear SVM problems.
    \\
    (ii). Algorithmically, {$L_{0/1}$-ADMM}, designed to solve linear $L_{0/1}$-SVM, requires data points $\mathcal{X}$ explicitly expressed in the iteration of $\textbf{w}$ and it is hampered by the fact that  in most cases  we  do not know  ${\Phi(\mathcal{X})}$ but their product dots.

To overcome these challenges, we  seek a new nonlinear model  suitable for the entire linear framework just reviewed, the following subsection will provide the essential basis to construct it.

\subsection{Kernel Method  and Representation Theorem}

In the kernel method,  the kernel function plays a central role and the representation theorem is also closely related to the kernel function, since it is the basic element to construct the Hilbert space in which the representation theorem holds.

 When failing to find a hyperplane as to the linearly inseparable data points $x_i \in {\mathcal{X}}(i \in {\mathbb{N}}_{m}, \mathcal{X} \subseteq \mathbb{R}^{n}) $, an ingenious solution is that we can transfer the original data set ${\mathcal{X}}$ into a new separable domain $\Phi({\mathcal{X}})$, here $\Phi({\mathcal{X}})$ is always a subset of dot product space ${\mathscr{H}}$ under the proper mapping $\Phi$. Since most nonlinear SVMs algorithms need to know only the dot products of the new domain $\Phi({\mathcal{X}})$ but not $\Phi({\mathcal{X}})$ themselves. Furthermore, ${\mathscr{H}}$ is  high dimensional, sometimes even with $\infty$, then  a new technique called kernel method is introduced to alleviate the computation burden in ${\mathscr{H}}$. The kernel function, playing a key role in this method, is defined as follows:
\begin{definition}
[Kernel function]\label{kernel f1} [25] For a given binary function
$k: {\mathcal{X}} \times {\mathcal{X}} \rightarrow \mathbb{R}$, $({\textbf{x}_i},{\textbf{x}_j})\mapsto k({\textbf{x}_i},{\textbf{x}_j})$,
there exists a map $\Phi$ from ${\mathcal{X}}$ to a dot product space ${\mathscr{H}}$, for all ${\textbf{x}_i},{\textbf{x}_j} \in {\mathcal{X}}$, satisfying
{$k({\textbf{x}_i},{\textbf{x}_j})=\langle \Phi({\textbf{x}_i}),\Phi({\textbf{x}_j})\rangle$\label{kernel de}}, then function $k(\cdot,\cdot)$ is usually called kernel function  and $\Phi(\cdot)$ is called its feature map.
\end{definition}
The most frequently used kernel function includes Polynomial kernel function [26] $k({\textbf{x}},{\textbf{x}}^{'})={\langle {\textbf{x}},{\textbf{x}}^{'}\rangle}^{d}$ $({d}\geq 1)$, Gaussian kernel function [27,28] $k({\textbf{x}},{\textbf{x}}^{'})=\exp(-\frac{\Vert {\textbf{x}}-{\textbf{x}}^{'} \Vert^{2}}{2{\sigma}^2} )$ $({\sigma} \textgreater 0)$, and Sigmoid kernel function [29] $k({\textbf{x}},{\textbf{x}}^{'})=\tanh(\beta{\langle {\textbf{x}},{\textbf{x}}^{'}\rangle}+\theta)$ $({\beta}\textgreater 1,{\theta}\textless 0)$, etc.

Next based on the kernel function, we introduce a special Hilbert space, the reproducing kernel Hilbert space (RKHS) [30], which is the completion of the inner space with all
${f(\cdot)}= {\sum_{i=1}^{m}} {\beta_i} k(\textbf{z}_{i},\cdot)\in {\mathbb{R}}^{{\mathcal{X}}}, {\beta_i \in {\mathbb{R}}}, \textbf{z}_i \in {\mathcal{X}},{\forall m\le n } \}$ as its elements. The corresponding dot product can be derived by $\langle k({\textbf{x}},{\cdot}),k({\textbf{y}},{\cdot})\rangle=k({\textbf{x}},{\textbf{y}})$,${\forall} {\textbf{x}},{\textbf{y}}\in {\mathcal{X}} $.

In the following part, we further review a representation theorem in RKHS, which is the cornerstone to construct our proposed nonlinear $L_{0/1}$-SVM.
In mathematics, there are many closely related variants of  the representation theorem. One of them is called Riesz representation theorem [31], which establishes an important connection between a Hilbert space and its continuous dual space.
Wahba [32] applied Riesz Representation theorem on RKHS and shows that the solutions of certain risk minimization problems, involving an empirical risk term and a quadratic regularizer, can be expressed as expansions in terms of  training examples.  Scholkopf et al. [24] further generalized the theorem to a larger class of regularizers and empirical risk terms, with the formulation as follows:

\begin{theorem}
[Nonparametric Representation Theorem]\label{NPtheorem}[24] Suppose we are
given a nonempty set ${\mathcal{X}}$, a positive definite real-valued kernel $k: {\mathcal{X}} \times {\mathcal{X}} \rightarrow {\mathbb{R}}$, a
training sample $(\textbf{x}_{i},y_{i})(i \in {\mathbb{N}}_{m}) \in  {\mathcal{X}} \times {\mathbb{R}} $, a strictly monotonically increasing
real-valued function $g$ on $[0,+\infty]$, an arbitrary cost function $c: ({\mathcal{X}} \times {{\mathbb{R}}^{2}})^{m} \rightarrow {{\mathbb{R}}\cup\{\infty\}}$, and a class of functions
\begin{center}
\begin{equation}
\displaystyle \mathcal{F}=\{f\in {\mathbb{R}}^{{\mathcal{X}}}\bigg|{f(\cdot)}= {\sum_{i=1}^{m}} {\beta_i} k({\textbf{z}}_{i},\cdot), {\beta_i \in {\mathbb{R}}}, {\textbf{z}}_i \in {\mathcal{X}},\left\|f\right\| \le \infty  \}.
\end{equation}
\end{center}
Here, $\left\|\cdot\right\|$ is the norm in the {RKHS}, ${\ H_k}$ associated with $k(\cdot,\cdot)$, i.e., for any ${\textbf{z}}_i \in {\mathcal{X}}$,
\begin{equation}
\label{rrkhs}
\left\|\sum_{i=1}^{\infty} {\beta_i} k({\textbf{z}}_{i},\cdot)\right\|^{2}={\sum_{i=1}^{\infty}} {\sum_{j=1}^{\infty}} {\beta_i} {\beta_j}k({\textbf{z}}_{i},{\textbf{z}}_{j}).
\end{equation}

Then any $f\in \mathcal{F} $ minimizing the regularized risk functional
\begin{eqnarray}
\label{rrkhs1}
c(({\textbf{x}}_1,y_1,f(x_1)),~\cdots~,({\textbf{x}}_m,y_m,f(x_m)))+ g(\left\|f\right\|)
\end{eqnarray}
admits a representation of the form \par
~~~~~~~~~~~~~~~~~~~~~~~~~~${f(\cdot)}= -\sum_{i=1}^m {\alpha_i} {y_i}k({\textbf{x}}_{i},\cdot),$\\
where $\alpha_i \in {\mathbb{R}}(i \in {\mathbb{N}}_{m})$ are coefficients of $f$ in the RKHS ${\ H_k}$.
\end{theorem}

\section{$L_{0/1}$-KSVM}\label{fnlsvm0}

In this section, we formally propose a nonlinear $L_{0/1}$-SVM
termed as  $L_{0/1}$-KSVM based on the kernel method and Theorem \ref{NPtheorem}, then its degeneracy property of the $L_{0/1}$-KSVM, the first-order optimality condition and solving algorithm are also explicitly explored.
\subsection{Modeling of $L_{0/1}$-KSVM}\label{fnlsvm000}
Given data set composed by  input and output pairs ${\mathcal{X}} \times {\mathbb{Y}}$ = $\{({\textbf{x}}_{1},{y}_{1}), \\ ({\textbf{x}}_{2},{y}_{2}),~\cdots~,({\textbf{x}}_{m},{y}_{m})\}$$ \subseteq {\mathbb{R}}^{n}\times \{{+1},{-1}\}$, the nonlinear kernel $L_{0/1}$-SVM classifier, termed as $L_{0/1}$-KSVM, is obtained by solving the following non-convex problem:
\begin{eqnarray} \label{gksvmnew}
&\underset{ \alpha \in {\mathbb{R}}^m,{\textbf{u}} \in {\mathbb{R}}^m}{\min} & \frac{1}{2} {\alpha}^T \widetilde{{\textbf{K}}}{\alpha} +C\|{{\textbf{u}}}_{+}\|_{0}\\ \nonumber
&\mbox{s.t.} &{{\textbf{u}}}- \widetilde{{\textbf{K}}}{\alpha} ={{\textbf{e}}},
\end{eqnarray}
where $\widetilde{{\textbf{K}}}$=${\textbf{V}}{\textbf{K}}\textbf{V}$, $\textbf{V}=\text{Diag}(\textbf{y})$, $\textbf{y}=(y_{1},y_{2},~\cdots~,y_{m})^{\top}$, the kernel matrix   ${{\textbf{K}}}=\left[k(\widetilde{\textbf{x}}_i,\widetilde{\textbf{x}}_j)\right]|_{i,j=1}^m$   is associated with the chosen kernel function $k(\cdot,\cdot)$
on modified input data set $\widetilde{\mathcal{X}}=\{{\widetilde{\textbf{x}}_{1}},{\widetilde{\textbf{x}}_{2}},~\cdots~,\widetilde{\textbf{x}}_{m}\}$ with ${\widetilde{\textbf{x}}_{i}}=({{\textbf{x}}_{i}}^{\top}, 1)^{\top}(i \in {\mathbb{N}}_{m})$.
The decision surface of the $L_{0/1}$-KSVM classifier is ${f}(\widetilde{\textbf{x}})=-\sum_{i=1}^m {\alpha_i}{{y}_i}k(\widetilde{\textbf{x}}_{i},\widetilde{\textbf{x}})$ for the transformed point $\widetilde{\textbf{x}}=({\textbf{x}}^{\top}, 1)^{\top}$ as to the test point ${\textbf{x}}\in {\mathbb{R}}^n$.

Next, we give some explanations about the modeling of the $L_{0/1}$-KSVM by Theorem \ref{NPtheorem}.
In \eqref{rrkhs1}, we are specifying that

\begin{center}
\begin{equation}
g(\left\|f\right\|)=\frac{1}{2}{\left\|f\right\|}^{2}
\end{equation}
\end{center}
and
\begin{center}
\begin{equation}
c(({\widetilde{\textbf{x}}_{1}},y_1,f({\widetilde{\textbf{x}}_{1}})),~\cdots~,(\widetilde{\textbf{x}}_{m},y_m,f({\widetilde{\textbf{x}}_{m}})))
= C\sum_{i=1}^{m}\ell_{0/1}(1-y_{i}f(\widetilde{\textbf{x}}_i))
\end{equation}
\end{center}
 Here, $C>0$ is the penalty parameter.\\
Therefore, the regularization problem for the nonlinear kernel SVM with 0-1 loss has the form:
\begin{eqnarray} \label{triformulate1new}
&\underset{ {f} \in {\ H_k}}{\min} & \frac{1}{2}{\|{f}\|}{^2}+C\sum_{i=1}^{m}\ell_{0/1}(1-y_{i}f(\widetilde{\textbf{x}}_i)).
\end{eqnarray}
Since  ${f(\cdot)}= -\sum_{i=1}^m {\alpha_i} {y_i}k(\widetilde{\textbf{x}}_{i},\cdot)$ is the solution of  \eqref{triformulate1new} due to Theorem \ref{NPtheorem},  we put it  back into (\ref{triformulate1new}) and apply the norm equality \eqref{rrkhs},  the final formulation \eqref{gksvmnew} of the $L_{0/1}$-KSVM is obtained.

 At the end of the subsection,  a significant proposition is introduced to reveal the degeneracy relationship between our newly-built nonlinear $L_{0/1}$-KSVM and the slightly modified edition of $L_{0/1}$-SVM.
\begin{prop}
[Degeneracy of $L_{0/1}$-KSVM to $L_{0/1}$-SVM]\label{Dtheorem}
The nonlinear $L_{0/1}$-KSVM \eqref{gksvmnew} with a linear kernel degenerates to  a slightly modified edition of the linear $L_{0/1}$-SVM \eqref{reformulate}, i.e.,  an extra regularization $\frac{1}{2} b^2$ is added in the objective function of the $L_{0/1}$-SVM.
\end{prop}
The proof of  Proposition \ref{Dtheorem} is in \ref{proof222}.
\subsection{ First-Order Optimality Condition}\label{sec:FOOC1}

In this subsection, the first-order optimality condition for the $L_{0/1}$-KSVM problem \eqref{gksvmnew} will be explored and the whole process is similar to the work completed in [20].
To proceed this, we first  define the pillar of the construction of the first-order optimality condition, the proximal stationary
point of $L_{0/1}$-KSVM:
\begin{definition}
[P-stationary point of $L_{0/1}$-KSVM]\label{pstationarypoint} For a given $C>0$, we say $({ \alpha^*}; \textbf{u}^* )$ is a proximal stationary (P-stationary) point of $L_{0/1}$-KSVM for a constant $\gamma>0$ such that
\be\label{aaalfi}
\left\{
\begin{array}{rll}   \textbf{u}^*-{\widetilde{\textbf{K}}} \alpha^*&={\textbf{e}},\\
\text{Prox}_{\gamma C\|({\cdot})_{+}\|_{0}}(\textbf{u}^*-\gamma{\alpha}^*)&=\textbf{u}^*,
\end{array}
\right.
\ee
where
\begin{equation}
\label{exp-proximal1LL0}
[\text{Prox}_{\gamma C\|({\cdot})_{+}\|_{0}}({ \textbf{z}^*})]_i=
\begin{cases}
0,& 0< z^*_i\leq\sqrt{2\gamma C},\\
z^*_i,& z^*_i>\sqrt{2\gamma C}~ \text{or}~z^*_i\leq0 .
\end{cases}
\end{equation}
Here, $\textbf{z}^*:=\textbf{u}^*-\gamma{\boldsymbol{\lambda}^*}$.
\end{definition}

Based on P-stationary point, the optimality condition of  $L_{0/1}$-KSVM is concluded as the following theorem:
\begin{theorem}\label{gol-p} For the $L_{0/1}$-KSVM problem \eqref{gksvmnew}, the following relations hold.

(i)~ Assume $\widetilde{\textbf{K}}$ is invertible. For a given $C>0$, if $(\alpha^*; {\textbf{u}^*})$ is a global minimizer of $L_{0/1}$-KSVM, then it is a P-stationary point of $L_{0/1}$-KSVM with $0<\gamma<  1/\lambda_{\max}(\widetilde{\textbf{K}}^{-1})=\lambda_{\min}(\widetilde{\textbf{K}})$, here $\lambda_{\min}(\widetilde{\textbf{K}})$ is the smallest eigenvalue of $\widetilde{\textbf{K}}$ .\label{gol-prox1}

(ii)~For a given $C>0$, if $(\alpha^*; {\textbf{u}^*})$ is a P-stationary point of $L_{0/1}$-KSVM with $\gamma>0$, then it is a local minimizer of $L_{0/1}$-KSVM.\label{gol-prox2}
\end{theorem}
The proof of  Theorem \ref{gol-p} is in \ref{proof22}.

In the following subsection, the classical algorithm ADMM is carried on to find a numerical solution to our nonlinear {$L_{0/1}$-SVM} problem,  written as   {$L'_{0/1}$-ADMM}. The SV of  $L_{0/1}$-KSVM  is defined and applied as select working set in   updating   all sub-problems, which leads to the termination condition satisfied faster.
\subsection{ $L'_{0/1}$-ADMM Algorithm via Selection of Working Set} \label{Fast}

\subsubsection{ $L_{0/1}$-KSVM Support Vector}\label{sec:alg11}
Suppose $(\alpha^*; {\textbf{u}^*})$ is a P-stationary  point, hence be a local minimizer of $L_{0/1}$-KSVM. From the second equation of \eqref{aaalfi}, we define the index set
\begin{eqnarray}\label{index}
T^{*} :=\left\{i \in {\mathbb{N}}_{m}:~\textbf{u}_i^{*}-\gamma{\alpha}_i^{*} \in (0,\sqrt{2\gamma C}]\right\},
\end{eqnarray}
and $\overline{T}^{*}: ={\mathbb{N}}_{m} \backslash T^{*}$.
Then ${\mathbb{N}}_{m}= T^{*}\cup\overline{T}^{*}$ and $T^{*} \cap \overline{T}^{*}=\emptyset$. By \eqref{exp-proximal1LL0}, we have
\begin{eqnarray*}
 \left[\begin{array}{c}
(\text{Prox}_{\gamma C \|(\cdot)_+\|_0}(\textbf{u}^{*}-\gamma{\alpha^{*}}))_{T^{*}}\\
(\text{Prox}_{\gamma C \|(\cdot)_+\|_0}(\textbf{u}^{*}-\gamma{\alpha^{*}}))_{\overline{T}^{*}}
\end{array}\right]
=\left[\begin{array}{c}
{\textbf{0}}_{T^{*}}\\
(\textbf{u}^{*}-\gamma{\alpha^{*}})_{\overline{T}^{*}}
\end{array}\right].
\end{eqnarray*}
This leads to
\begin{eqnarray}\label{proxm1}
 {\textbf{u}^{*}} =\text{Prox}_{\gamma C \|(\cdot)_+\|_0}(\textbf{u}^{*}-\gamma{\alpha^{*}})~~\Longleftrightarrow~~
\begin{bmatrix}
\textbf{u}^{*}_{T^{*}}\\
\alpha^{*}_{\overline{T}^{*}}
\end{bmatrix}={\textbf{0}}.
 \end{eqnarray}
Then from definition of $T^{*}$ and \eqref{proxm1}, we have
$$\textbf{u}^{*}_{i} -\gamma {\alpha}_i^{*}= -\gamma{\alpha}_i^{*} \in (0,\sqrt{2\gamma C}], i \in T^{*}$$
i.e.,
$${\alpha}_i^{*} \in [-\sqrt{2 C/\gamma},0), i \in T^{*}. $$
Combining with ${\alpha}_i^{*}=0$  $(i \in \overline{T}^{*})$ in \eqref{proxm1} yields
\begin{eqnarray}\label{lambdaSV}
{\alpha}_i^{*}
\begin{cases}
\in[- \sqrt{2 C/\gamma},0),& \text{for} ~~i \in T^{*},\\
=0,& \text{for}~~ i \in \overline{T}^{*}.
\end{cases}
\end{eqnarray}
Taking \eqref{lambdaSV} into the function of decision surface of $L_{0/1}$-KSVM,  we derive\par
~~~~~~~~~${f}({\widetilde{\textbf{x}}})=-\sum_{i=1}^m {\alpha_i}^{*}{y_i}k({\widetilde{\textbf{x}}}_{i},\widetilde{\textbf{x}})=-\underset{i \in T^{*}}{\sum} {\alpha_i}^{*}{y_i}k({\widetilde{\textbf{x}}}_{i},{\widetilde{\textbf{x}}}).$\\
Following the concept of SV in [1], $\left\{{\widetilde{\textbf{x}}}_{i}: i \in T^{*}\right\}$ or $\left\{{\textbf{x}}_{i}: i \in T^{*}\right\}$ are the SVs of the {$L_{0/1}$-KSVM}, here $T^{*}$ becomes the index set of SVs. Again, from $u^{*}_{i}=1-y_{i}f({{\widetilde{\textbf{x}}}_{i}})=0$ for $i \in T^{*}$, we obtain
\begin{eqnarray}\label{proxm1222}
f({{\widetilde{\textbf{x}}}_{i}})=\pm1.
%\langle\bfw^{*},\bfx_{i}\rangle+b^{*}\neq\pm1, ~\text{for}~ i \in \overline{T}^{*}.
 \end{eqnarray}
That is to say, all the SVs of {$L_{0/1}$-KSVM} lie on the support surfaces $f({{\widetilde{\textbf{x}}}_{i}})=-\underset{i \in T^{*}}{\sum} {\alpha_i}^{*}{y_i}k({\widetilde{\textbf{x}}}_{i},{\widetilde{\textbf{x}}})=\pm1$. As far as we know, only the hard-margin SVM  and $L_{0/1}$-SVM have such a property. This geometric  property shows  that the SVs of {$L_{0/1}$-KSVM} arrange orderly and probably sparsely. Moreover, this elegant theoretical conclusion also shows its power  on the subsequent designing of an efficient algorithm for {$L_{0/1}$-KSVM}.

\subsubsection{ $L'_{0/1}$-ADMM Algorithm} \label{Fast1}

First, to the best of our knowledge, the closed-form solution of   $L_{0/1}$-KSVM is hard to trace due to its non-convexity, although its existence can be verified by using similar skill in [20]. In this subsection, the ADMM algorithm  is applied to  the augmented problem of \eqref{gksvmnew} to find its numerical solution:
 \begin{eqnarray}
L_{\sigma}(\alpha;\textbf{u};{\boldsymbol{\lambda}})=\frac{1}{2} {\alpha}^T \widetilde{\textbf{K}}{\alpha} +C\|\textbf{u}_{+}\|_{0}+\langle{\boldsymbol{\lambda}},\textbf{u}-\textbf{e}-\widetilde{\textbf{K}}\alpha\rangle \\ \nonumber +\frac{\sigma}{2}\|\textbf{u}-\textbf{e}-\widetilde{\textbf{K}}\alpha\|^{2},~~~~~~~~~~~~~~~~~~~~~~~~~~~~~~~
\end{eqnarray}
where ${{\boldsymbol{\lambda}}}$ is the Lagrangian multiplier and $\sigma>0$ is the penalty parameter. Given the $k$th iteration $(\alpha^k;{\boldsymbol{\lambda}}^k)$, the algorithm framework can be described as
\begin{eqnarray}\label{ADMM}
\begin{cases}
{\textbf{u}}^{k+1}&=\underset{{\textbf{u}}\in\mathbb{R}^m}{\rm argmin}~L_{\sigma}( \alpha ^{k},{\textbf{u}},{\boldsymbol{\lambda}}^{k}),\\
\alpha ^{k+1}&=\underset{\alpha\in\mathbb{R}^n}{\rm argmin}~L_{\sigma}( \alpha ,{\textbf{u}}^{k+1},{\boldsymbol{\lambda}}^{k})+{\frac{\sigma}{2}}\|\alpha-\alpha^k\|^2_{D_k},\\
{\boldsymbol{\lambda}}^{k+1}&={\boldsymbol{\lambda}}^{k}+\eta\sigma(\textbf{u}^{k+1}-\textbf{e}-\widetilde{\bf{K}}\alpha^{k+1}),
\end{cases}
\end{eqnarray}
%\noindent {\bf (i) Updating ${T}_{k}$.} By (\ref{index}), we denote
%\begin{eqnarray}\label{zk1}
%T_{k}= \left\{i\in \N_m:\left[\text{prox}_{ \gamma C L_{0/1}}({\bfe}-A \bfw ^{k}-b^{k}\bfy- \frac{\bm{\lambda}^{k}}{\sigma})\right]_i=0\right\}.
%\end{eqnarray}
where $\eta>0$ is referred as the dual step-size. Here,
$$\|\alpha-\alpha^k\|^2_{D_k}=\langle\alpha-\alpha^k,  D_k(\alpha-\alpha^k)\rangle$$
is the so-called proximal term, $D_k\in\mathbb{R}^{n\times n}$ is a symmetric matrix properly chosen to guarantee the convexity of ${\alpha}$-subproblem  of \eqref{ADMM}.\\
\noindent {\bf (i) Updating $\textbf{u}^{k+1}$}: The $\textbf{u}$-subproblem in \eqref{ADMM} is equivalent to the following problem
\begin{eqnarray*}\label{prox_u0}\nonumber
&&\textbf{u}^{k+1}\\\nonumber
&=&\underset{\textbf{u}\in\mathbb{R}^{m}}{\rm argmin}~C\|\textbf{u}_{+}\|_{0}+\langle{\boldsymbol{\lambda}}^{k},\textbf{u}\rangle+\frac{\sigma}{2}\|\textbf{u}-\textbf{e}-\widetilde{\bf{K}}\alpha^{k}\|^{2}\\ \nonumber
%&=&\underset{\bfu\in \R^{m}}{\rm argmin}~C\|\bfu_{+}\|_{0}+\frac{\sigma}{2}\|\bfu-(\bfe-A\bfw^{k}-b^{k}\bfy-{\bm{\lambda}^{k}}/{\sigma})\|^{2} \nonumber\\
&=&\underset{\textbf{u}\in\mathbb{R}^{m}}{\rm argmin}~C\|\textbf{u}_{+}\|_{0}+\frac{\sigma}{2}\|\textbf{u}-\textbf{z}^k\|^{2} \\
&=&\text{Prox}_{\frac{C}{\sigma}\|\textbf(\cdot)_{+}\|_{0}}(\textbf{z}^k),
\end{eqnarray*}
where the third equation is obtained from the definition of proximal operator with $\gamma=1/\sigma$ and $\textbf{z}^k:=\textbf{e}+\widetilde{\bf{K}}{\alpha^{k}}-{{\boldsymbol{\lambda}}}^{k}/{\sigma}$. Through defining the working set
\begin{eqnarray}
\label{prox_T}T_k :=\left\{i\in \mathbb{N}_m:~\textbf{z}^k_i \in (0,\sqrt{2C/\sigma}]\right\}
\end{eqnarray}
and complementary set $\overline{T}_k:={\mathbb{N}}_m \backslash T_k,$   we obtain
\begin{eqnarray}
\label{prox_u}{\textbf{u}}^{k+1}_{T_k}= {\bf0},~~~
{\textbf{u}}^{k+1}_{\overline{T}_k}= \textbf{z}^k_{\overline{T}_k}.
\end{eqnarray}

\noindent {\bf (ii) Updating $\alpha^{k+1}$.}
\begin{eqnarray}\label{dk}
D_{k}= -\widetilde{\bf{K}}_{\overline{T}_k}^\top\widetilde{\bf{K}}_{\overline{T}_k}
\end{eqnarray}
is introduced during the updating process, thus we derive the ${\alpha}$-subproblem of \eqref{ADMM} as
\begin{eqnarray}\label{wk1}
\alpha^{k+1}= \underset{\alpha \in\mathbb{R}^{m}}{\rm argmin}\frac{1}{2} {\alpha}^T \widetilde{\bf{K}}{\alpha} + \frac{\sigma}{2}\| \alpha - \alpha^k\|^{2}_{-\widetilde{\bf{K}}_{\overline{T}_k}^\top \widetilde{\bf{K}}_{\overline{T}_k}} \nonumber\\
-\langle{\boldsymbol{\lambda}}^{k},\widetilde{\bf{K}}\alpha\rangle+\frac{\sigma}{2}\|\textbf{u}^{k+1}-\textbf{e}-\widetilde{\bf{K}}\alpha\|^{2}~~
\end{eqnarray}
and it can be simplified as
\begin{eqnarray}\label{euqation-w}
(\widetilde{\bf{K}} +\sigma{\widetilde{\bf{K}}_{T_k}}^\top{\widetilde{\bf{K}}_{T_k}}) \alpha={ \sigma} {\widetilde{\bf{K}}_{T_k}}^\top \textbf{v}_{T_k}^k ,
\end{eqnarray}
where $\textbf{v}^k:={\textbf{u}}^{k+1}-{\textbf{e}}+{{\boldsymbol{\lambda}}}^{k}/\sigma$, then we update $ \alpha^{k+1}$ by applying Conjugate Gradient(CG) method [33] on \eqref{euqation-w} for efficiency.

\noindent {\bf (iii) Updating $\boldsymbol{\lambda}^{k+1}$.} Let $\boldsymbol{\lambda}^{k+1}_{\overline{T}_k}= {\bf 0}$, then based on \eqref{ADMM}, we have
\begin{eqnarray}\label{lambda-k}
\begin{cases}
\boldsymbol{\lambda}^{k+1}_{T_k}=&\boldsymbol{\lambda}^{k}_{T_k}+\eta\sigma\boldsymbol{\varpi}^{k+1}_{T_k},\\
\boldsymbol{\lambda}^{k+1}_{\overline{T}_k}=& {\bf 0},
\end{cases}
\end{eqnarray}
where $\boldsymbol{\varpi}^{k+1}:=\textbf{u}^{k+1}-\textbf{e}-\widetilde{\bf{K}}\alpha^{k+1}$.

Summarizing  \eqref{prox_T},\eqref{prox_u}, the solution of \eqref{euqation-w} and \eqref{lambda-k}, we obtain Algorithm \ref{Alg-SQREDM}, which is called the method { $L'_{0/1}$}-ADMM,  an abbreviation for {$L_{0/1}$-KSVM} solved by  ADMM.

\begin{algorithm} %\label{Alg-SQREDM}
	\caption{:  $L'_{0/1}$-ADMM for solving $L_{0/1}$-KSVM \eqref{gksvmnew} }
	\label{Alg-SQREDM}
\begin{algorithmic}
	\STATE{Initialize ($ \alpha^0;{\textbf{u}}^0;{\boldsymbol{\lambda}}^0$). Choose parameters  $C,\eta,\sigma,K>0$ and set $k=0$.}
 \WHILE{The halting condition does not hold and $k\leq K$}
 	\STATE{Update $T_k$ as in \eqref{prox_T}.}
 \STATE{Update $\textbf{u}^{k+1}$ by \eqref{prox_u}.}
\STATE{Update $\alpha^{k+1}$ by CG method on \eqref{euqation-w}.}
\STATE{Update ${\boldsymbol{\lambda}}^{k+1}$ by \eqref{lambda-k}.}
\STATE{Set $k=k+1$.}
 \ENDWHILE
 \RETURN the solution ($ \textbf{u} ^k,\alpha^k$) to \eqref{gksvmnew}.
\end{algorithmic}
\end{algorithm}
\subsubsection{$L'_{0/1}$-ADMM Convergence and Complexity Analysis}\label{calculation}
First, considering that directly solving the SVM with 0-1 loss  is a NP-hard problem, we admit that there are  limited materials  available to help us   analyze the solution system. Moreover,
the ADMM algorithm, that we use to calculate the $L_{0/1}$-KSVM, has intrinsic flaw in analysis of convergence because  it is designed to have a multi-step and multi-factor structure compared to other popular algorithms such as gradient algorithms. Further, $T_k$, the select working
set in the updating of all sub-problems in $L'_{0/1}$-ADMM,  varies in each iteration, which undoubtedly aggravates the instability of the whole computational process.
  Even worse,   this algorithm now has to address  is  a non-convex and discontinuous problem unlike  optimization problems with a general class of friendlier non-convex losses, such as the tasks terminated in [34-38].
Nevertheless, we arrive at a convergence conclusion similar to the linear form in [20] for the sequence generated by ADMM algorithm:
\begin{theorem}\label{convergence}Suppose that $( \alpha ^*;{\textbf{u}}^*;{\boldsymbol{\lambda}}^*)$ is the limit point of the sequence  $\{( \alpha^k;{\textbf{u}}^k;{\boldsymbol{\lambda}}^k)\}$ generated by $L'_{0/1}$-ADMM, then $({ \alpha^* }; {\textbf{u}}^*;{\boldsymbol{\lambda}}^*)$  is a local optimal solution of $L_{0/1}$-KSVM.
\end{theorem}
The proof of the  Theorem \ref{convergence} could be found in \ref{conver1}.

Next we present the computational complexity analysis of $L'_{0/1}$-ADMM.
\begin{itemize}
\item Updating $T_k$ by \eqref{prox_T} needs the complexity $\mathcal{O}(m)$.
\item The main term involved in computing $\textbf{u}^{k+1}$ by \eqref{prox_u} is the kernel matrix ${\bf{K}}$, taking the complexity about $\mathcal{O}({m^2}n)$, $n$ is the dimension of data points $\textbf{x}_{i}  (i \in {\mathbb{N}}_{m})$.
\item To update $\alpha^{k+1}$, calculating $${\widetilde{\bf{K}}_{T_k}}^\top{\widetilde{\bf{K}}_{T_k}}$$ will cost $\mathcal{O}(m^2|T_k|)$ as its computational complexities. Moreover, we need $\mathcal{O}({m^2}q)$ [39] to compute \eqref{euqation-w} by CG method, here $q$ is the number of distinct eigenvalues of $\widetilde{\bf{K}} +\sigma{\widetilde{\bf{K}}_{T_k}}^\top{\widetilde{\bf{K}}_{T_k}}$. Therefore, the  complexity to update $\alpha^{k+1}$ in each step is $$\mathcal{O}(m^2\max\{|T_k|,q\}).$$
\item Similar to updating $\textbf{u}^{k+1}$, $\widetilde{\bf{K}}\alpha^{k+1}$ is the most expensive computation in \eqref{lambda-k} to derive ${\boldsymbol{\lambda}}^{k+1}$. Its complexity is $\mathcal{O}(m^2)$ supposed we have calculated ${\bf{K}}$ firstly.
\end{itemize}
Overall, the whole computational complexity in each step of  $L'_{0/1}$-ADMM in Algorithm \ref{Alg-SQREDM} is
$$\mathcal{O}(m^2\max\{|T_k|,q,n\}).$$

Obviously, the handling of the kernel matrix $\widetilde{\bf{K}}$  leads the major computational burden, we can call it  the curse of kernelization [40], which continues to be an open and big challenge in kernel SVM training problems.

\section{ Numerical experiments}\label{fnlsvm088}
The experiments related to our proposed algorithm $L'_{0/1}$-ADMM on $L_{0/1}$-KSVM (abbreviated as $L'_{0/1}$ in the following tables), are divided into two parts, corresponding to two  entities as a performance contrast: the linear $L_{0/1}$-SVM and the other six leading nonlinear SVM classifiers. 10 UCI data sets, whose detailed information is listed in Table \ref{Table-LU-20}, are selected to conduct all experiments
using MATLAB (2019a) on a laptop with 32GB of memory and Inter Core i7 2.7Ghz CPU. In addition,  all features in each data set are scaled to $[-1,1]$ and Gaussian kernel is chosen in all experiments.
\begin{table*}[htbp]
\begin{center}
\renewcommand\tabcolsep{1.0pt}
\caption{Descriptions of 10 UCI data sets.}
\resizebox{3.5in}{!}
{
\begin{tabular}{lccccclccccc}
\toprule  {\multirow{1}{*}{Data sets}}&  ~~~{Numbers($m$)} & ~~~ {Features($n$)}   &~~~Rate($\pm$)\\
\midrule
 Breast (\texttt{bre})     &699  &9&34/66\\
			Echo(\texttt{ech})    & 131   & 10 &33/67 \\
            Heartstatlog (\texttt{hea})     &270           &13&56/44\\
			Housevotes(\texttt{hou})    & 435         &    16 &61/39 \\
            Hypothyroid (\texttt{hyp})     &3163          &25&5/95\\
			Monk3(\texttt{mon})    & 432         &    6&50/50  \\
            PimaIndian (\texttt{pim})     &768           &8&35/65\\
			TwoNorm (\texttt{two})    & 7400  &  20&50/50   \\
			Waveform (\texttt{wav})     &5000           &21&33/67\\
			WDBC (\texttt{wdb})    & 198  &  34&37/63   \\
\bottomrule
\end{tabular}
}
\label{Table-LU-20}
\end{center}
\end{table*}
 Theorem \ref{gol-p} allows us to take the P-stationary point as a stopping criterion, then we can terminate our algorithm if the point ($ \alpha ^k;{\textbf{u}}^k;{\boldsymbol{\lambda}}^k$) closely satisfies the conditions in \eqref{aaalfi}, namely, $$\max\{\theta^k_1,\theta^k_2\}<\texttt{tol},$$  where \texttt{tol} is the tolerance level  and
\begin{eqnarray*}
\theta^k_1&:=&\frac{\|{\textbf{u}}^k-{\textbf{e}}-\widetilde{\bf{K}}\alpha^{k}\|}{\sqrt{m}},\\
\theta^k_2&:=&\frac{\|{\textbf{u}}^k-\text{Prox}_{\gamma C\|({\cdot})_{+}\|_{0}}({\textbf{u}}^k-\gamma{\alpha}^{k})\|}{1+\|{\textbf{u}}^k\|}.
\end{eqnarray*}\\
For our evaluation of classification performances is based on the method of 10-fold cross validation, that is, each data set is randomly split  into ten parts, one of which is used for testing and the remaining nine parts is for training. Therefore, we choose the mean accuracy $({\mACC})$, the mean number of support vectors $({\mNSV} )$  and the mean CPU time $({\mCPU})$ as the criteria of performance of all models. Let $\{(\textbf{x}_j^{\rm test}, y_j^{\rm test})\}_{j=1}^{m_{t}}$ be $m_{t}$ test samples, the testing accuracy is defined by
$${\texttt{ACC}}:=1-\frac{1}{2m_t}\sum_{j=1}^{m_t} \Big|{\rm sign}(-\sum_{i \in T^{*}} {\alpha_i} {y_i}k({\widetilde{\textbf{x}}_j}^{\rm test},\widetilde{\textbf{x}}_i))-y_j^{\rm test}\Big|,$$  where $T^{*}$ is the SVs set of $L_{0/1}$-KSVM and {  $\{{\widetilde{\textbf{x}}_j}^{\rm test}\}_{j=1}^{m_{t}}$}
={  $\{(\textbf{x}_j^{\rm test}, 1)\}_{j=1}^{m_{t}}$}.

\subsection{Comparisons between linear $L_{0/1}$-SVM and  $L_{0/1}$-KSVM}

 The optimal parameters $C$ and $\sigma$ in two ADMM algorithms are  determined by the standard 10-fold cross validation  in the same range of\\  $\{2^{-8},2^{-7},\cdots,2^{8}\}$.  In addition, as to the other parameters options in $L_{0/1}$-ADMM, we set $\eta=1.618$ and  choose  $\textbf{w}^0 = -\textbf{e}/100$,  $b=0$, ${\boldsymbol{\lambda}}^0={\bf0}$ as  initial points, the maximum iteration number is $K=10^3$ and the tolerance level \texttt{tol}$=10^{-3}$.
 In the algorithm  $L'_{0/1}$-ADMM, $\eta=1$, $ \alpha ^0=0.01\times{\textbf{e}}$ and ${\boldsymbol{\lambda}}^0={\bf0}$, the maximum iteration number and the tolerance level are $K=10^2$,  \texttt{tol}$=10^{-3}$.
The comparison results of linear and nonlinear $L_{0/1}$-SVM can be found in Table \ref{Table-LU-3}. We conclude that the $L_{0/1}$-KSVM performs significantly better than its linear formulation in terms of $\mACC$ and $\mNSV$ for the majority of 10 data sets,  with  bold numbers indicating superiority. All data sets except  $\texttt{wdb}$ achieve a higher $\mACC$ for our proposed nonlinear model, the data set ${\texttt{mon}}$ increases its accuracy increment by the surprising 25$\%$, $\texttt{wav}$   reduces its error rate by almost 4.2$\%$. As to the $\mNSV$, $L_{0/1}$-KSVM also greatly outperforms its linear peer besides $\texttt{hea}$. Especially, the number of SVs occurring in the nonlinear formulation is  less than a fifth to its linear opponent in data sets $\texttt{hyp}$, $\texttt{two}$, $\texttt{ech}$.
The stronger sparsity is a little surprise to us considering that most convex nonlinear SVMs brings more SVs than its corresponding linear forms. We assume that higher accuracy in $L_{0/1}$-KSVM and the special geometrical distribution of SVs in 0-1 loss SVM are responsible for this unusual phenomenon.
 However, the curse of kernelization hinders the computation inefficiency for the $L'_{0/1}$-ADMM, which costs much more time than $L_{0/1}$-ADMM.

\begin{table*}[htbp]
\begin{center}
\renewcommand\tabcolsep{1.0pt}
\caption{The comparison of  $L_{0/1}$-SVM  and  $L_{0/1}$-KSVM on 10 UCI data sets.}
\resizebox{4.5in}{!}
{
\begin{tabular}{lccccclccccc}
\toprule  {\multirow{1}{*}{Data set}}&  ~~~~~{Catagory} & ~~~~~ {\mACC(\%)}   &~~~~~\mNSV&~~~~~{\mCPU(seconds)}\\
\midrule
 \multirow{2}{*}{\texttt{bre}}
&$L_{0/1}$-SVM& {0.9568} & {15.55}& {\textbf{0.0465}}    \\
\cline{2-5}
&$L_{0/1}$-KSVM& {\textbf{0.9677}} & {\textbf{9.82}}& {0.6453} \\
\cline{2-5}

\hline

\multirow{2}{*}{\texttt{ech}}
&$L_{0/1}$-SVM& {0.9042} & {23.35}& {\textbf{ 0.0381}} \\
\cline{2-5}
&$L_{0/1}$-KSVM&{\textbf{ 0.9186}} & {\textbf{7.76 }}& {0.0601} \\
\cline{2-5}
\hline
\multirow{2}{*}{\texttt{hea}}
&$L_{0/1}$-SVM& {0.8252} & {\textbf{18.67 }}& {\textbf{0.0349}} \\
\cline{2-5}
&$L_{0/1}$-KSVM& {\textbf{0.8374}} & {26.76} & {0.0912} \\
\cline{2-5}

\hline
\multirow{2}{*}{\texttt{hou}}
&$L_{0/1}$-SVM& {0.9458} & {27.85}& {\textbf{0.0297 }} \\
\cline{2-5}
&$L_{0/1}$-KSVM& {\textbf{0.9584}} & {\textbf{23.93 }}& {0.2323} \\
\cline{2-5}

\hline
\multirow{2}{*}{\texttt{hyp}}
&$L_{0/1}$-SVM& {0.9796} & {139.93} & {\textbf{0.1253 }}    \\
\cline{2-5}
&$L_{0/1}$-KSVM& {\textbf{0.9816 }} & {\textbf{ 25.07}} & {20.7255}  \\
\cline{2-5}
\hline
\multirow{2}{*}{\texttt{mon}}
&$L_{0/1}$-SVM& {0.7272} & {83.55} & {\textbf{0.0178 }}    \\
\cline{2-5}
&$L_{0/1}$-KSVM& {\textbf{0.9723 }} & {\textbf{78.73 }} & {0.2546}  \\
\cline{2-5}
\hline
\multirow{2}{*}{\texttt{pim}}
&$L_{0/1}$-SVM& {0.7647} & {267.18} & {\textbf{0.1710}}    \\
\cline{2-5}
&$L_{0/1}$-KSVM& {\textbf{0.7723 }} & {\textbf{176.54 }} & {0.8658}  \\
\cline{2-5}
\hline
\multirow{2}{*}{\texttt{two}}
&$L_{0/1}$-SVM& {0.9770} & {309.77} & {\textbf{0.3979 }}    \\
\cline{2-5}
&$L_{0/1}$-KSVM& {\textbf{0.9776 }} & {\textbf{16.81 }} & {116.4893}  \\
\cline{2-5}

\hline
\multirow{2}{*}{\texttt{wav}}
&$L_{0/1}$-SVM& {0.8539} & {740.53} & {\textbf{0.3969 }}    \\
\cline{2-5}
&$L_{0/1}$-KSVM& {\textbf{0.8955 }} & {\textbf{104.85 }} & {59.0364}  \\
\cline{2-5}

\hline
\multirow{2}{*}{\texttt{wdb}}
&$L_{0/1}$-SVM& {\textbf{0.9754 }} & {33.02} & {\textbf{0.0725 }}    \\
\cline{2-5}
&$L_{0/1}$-KSVM& {0.9744} & {\textbf{11.73 }} & {0.3385}  \\
\bottomrule
\end{tabular}
}
\label{Table-LU-3}
\end{center}
\end{table*}
\subsection{Comparisons between $L_{0/1}$-KSVM and benchmark nonlinear SVM classifiers}

 Six  widely-used nonlinear SVM classifiers are introduced to compare the performance with $L_{0/1}$-KSVM on 10 binary UCI data sets in Table \ref{Table-LU-20}.

{\bf  Benchmark classifiers.} Six SVM nonlinear classifiers  are introduced to make performance comparisons. All their parameters are also optimized to maximize the accuracy by 10-fold cross validation.

\begin{itemize}
  \item[\libsvc] SVM with hinge  soft-margin loss is implemented by LibSVM [41], where the parameter $C$ is selected from the set $\Omega:=\{2^{-8},2^{-7},\cdots,2^{8}\}$.

  \item[\lssvc] SVM with square  soft-margin  loss [9] is implemented by LS-SVMlab [42], where the parameter $C$ is selected from the range $\Omega$.

  \item[\l2svc] SVM with the squared hinge loss [1]  is solved by quadratic programming, where the parameter $C$ is selected from the range $\Omega$.

  \item[\ramp] SVM with ramp  soft-margin  loss can be tackled by employing the \texttt{CCCP} [13], where the parameter $C$  is selected from $\Omega$ and the $s$ is selected from $\{-1,-0.5,0\}$.
  \item[\rsvc] SVM with  the rescaled hinge loss [15] is addressed by employing the  HQ optimization method, where the parameter $C$ is selected from  $\Omega$  and the parameter $\eta$  is  selected from $\{0.2 , 0.5 , 1 , 2 , 3\}$.
  \item[\rshsvc] SVM with  non-convex robust and smooth soft-margin  loss [16] can be solved by employing the  iteratively reweighted algorithm with QP, where the parameter $C$ is selected from  $\Omega$  and the parameter $\sigma$  is selected from $\{0.4,0.6,0.8\}$.

\end{itemize}

\begin{example}[Real data without outliers]\label{ex:real-data-no-outlier}\end{example}

 The results compared with other nonlinear SVM classifiers on 10 UCI data sets without outliers are recorded  in Table \ref{Table-LU-11}.  The $L'_{0/1}$-ADMM algorithm, designed for $L_{0/1}$-KSVM, shows the dominant advantage  of sparsity while retaining   much of its generalization capability.
In the $\texttt{ech}$, $\texttt{hea}$ and $\texttt{hyp}$ data sets, $L_{0/1}$-KSVM exhibits the highest accuracy and strongest sparsity among all seven nonlinear classifiers. In particular, when we observe the results of  the three largest data sets $\texttt{hyp}$, $\texttt{two}$ and $\texttt{wav}$,
  the ratio of SVs for $L_{0/1}$-KSVM  is only 14$\%$, 5$\%$ and 11$\%$ to the algorithm ${\rshsvc}$, the  second best classifier on the index of sparsity scale. Moreover, we   found that the number of SVs in $L_{0/1}$-KSVM does not increase linearly with the size of training set, a characteristic usually found on convex nonlinear classifiers $\libsvc$, $\lssvc$, $\l2svc$ and this is  essential for a predictor when applied in relatively large data sets.
Meanwhile, the experiment results  show that non-convex SVMs perform better overall than convex peers at sparsity level. Furthermore, $L_{0/1}$-KSVM  also obtains a comparable decent outcome on accuracy $(\mACC)$ with the other non-convex classifiers ${\rsvc}$, ${\rshsvc}$ in all 7 algorithms. Nevertheless, these 3 non-convex classifiers, plus ${\ramp}$, pay at a big price of computing complexity, even our designed $L'_{0/1}$-ADMM exhausts less time than ${\rsvc}$, ${\rshsvc}$.
\begin{table*}[htbp]
\begin{center}
\renewcommand\tabcolsep{2.5pt}
\caption{Comparisons of seven classifiers on 10 UCI data sets.}
\resizebox{4.5in}{!}
{
\begin{tabular}{lccccclccccc}
\toprule
&\multicolumn{6}{c}{\mACC~(\%)}\\
        \text{Name} & $L'_{0/1}$ &{\libsvc}  &{\lssvc}  &{\l2svc} &{\ramp}&{\rsvc}&{\rshsvc} \\\hline
		\texttt{bre}  	& 96.77&96.76 	&	96.65	&	96.72	&	96.83	&	 \textbf{96.94}& 96.75 \\
       \texttt{ech}  	&	\textbf{91.87}	&	90.13	&	89.54	&	90.97	&	90.94&91.59 &	91.26	\\
	    \texttt{hea}  	&	\textbf{83.74}	&	83.26	&	82.56	&	83.56	&	82.85& 82.67 &	82.93	\\
     	\texttt{hou}	&	 95.84 	&96.24 	&	96.06 	&	96.00 		&	 95.63& \textbf{96.29} & 96.14		\\
		\texttt{hyp}	&	\textbf{98.16}	&	97.81	&	98.11	&	97.97	&	98.14& 97.65&97.74	\\
	    \texttt{mon}	&	97.23	&	\textbf{99.86}	&	97.39	&	99.05	&	96.23&\textbf{99.86}&99.65\\
        \texttt{pim}	&77.23	&	77.27	&	76.89	&	77.18	&	\textbf{77.59}&76.72 &77.27		\\		
     	\texttt{two}	&97.76	&	97.83	&	97.80 	&	97.83	&	97.67&\textbf{97.88} & 97.86		\\
		\texttt{wav}  	&89.55	&	89.97	&	89.97	&	89.98	&	89.50& 89.94&	\textbf{89.99}	\\
     	\texttt{wdb} 	&97.44	&	97.84	& 	\textbf{97.99}  &	97.95&	97.89	& 97.66&	97.91	\\\hline
&\multicolumn{5}{c}{\mNSV} \\
			 \text{Name} &$L'_{0/1}$ &{\libsvc}  &{\lssvc}  &{\l2svc} &{\ramp}&{\rsvc}&{\rshsvc} \\\hline
		\texttt{bre}  	& \textbf{9.82}&55.06 	&	629.10	&	153.98	&	34.48	&	 332.15& 73.08 \\
       \texttt{ech}  	&	\textbf{7.76}	&	36.46	&	117.90	&	70.81	&	27.45&90.55 &	30.35	\\
	    \texttt{hea}  	&	\textbf{26.76}	&	112.11	&	243.00	&	215.94	&	73.86& 216.82 &	92.67	\\
     	\texttt{hou}	&	 \textbf{23.93}&47.07 	&	391.50 	&	97.26		&	 64.42& 94.13& 75.48\\
		\texttt{hyp}	&	\textbf{25.07}	&	475.46	&	2846.70	&	468.08	&	346.80& 423.06&180.79	\\
	    \texttt{mon}	&	\textbf{78.73}	&122.04	&388.80	&	160.79	&	271.31&117.88&151.99\\
        \texttt{pim}	&\textbf{176.54}	&	425.64	&	691.20	&	624.60	&335.63&415.53 &341.42		\\		
     	\texttt{two}	&\textbf{16.81}&	465.82	&	5920.00 	&	943.27	&	878.53&407.26 &386.21		\\
		\texttt{wav}  	&\textbf{104.85}&	1356.36	&	4000.00	&	2892.71	&	1088.04& 1143.17&997.03	\\
     	\texttt{wdb} 	&	\textbf{11.73}&	69.06	& 512.10	&	126.10&	108.00	& 76.71&	76.14	\\\hline
&\multicolumn{5}{c}{\mCPU~(seconds)} \\
			 \text{Name} &$L'_{0/1}$     &{\libsvc}  &{\lssvc}   &{\l2svc}  &{\ramp} &{\rsvc} &{\rshsvc} \\\hline
		\texttt{bre}  	& 0.6453&\textbf{0.0041} 	&	0.0259	&	0.0664	&	0.1168	&	0.6705& 1.5803 \\
       \texttt{ech}  	&	0.0601&\textbf{0.0008}		&	0.0120	&	0.0043	&	0.0093&0.0292 &	0.2339	\\
	    \texttt{hea}  	&	0.0912&\textbf{0.0029}		&	0.0158	&	0.0104	&	0.0241& 0.0359 &	0.7212	\\
     	\texttt{hou}	&	0.2323&\textbf{0.0029}		&	0.0215	&	0.0275	&	0.0654& 0.1995 &	1.1385	\\
		\texttt{hyp}	&20.726	&\textbf{0.1288}		&	0.5271	&	3.2342	&56.852& 28.127&187.09	\\
	    \texttt{mon}	&	0.2546	&0.0172	&\textbf{0.0146}	&	0.0218	&	0.0192&0.0463&0.3023\\
        \texttt{pim}	&0.8658	&	\textbf{0.0228}	&	0.0307	&	0.0767	&0.3305&0.4779 &3.1791		\\		
     	\texttt{two}	&116.49	&	\textbf{0.1288}	&	1.8732	&	27.457	&	52.085&272.09 & 1355.4		\\
		\texttt{wav}  	&59.036	&	\textbf{0.2172}	&	0.7537	&	9.4656	&37.947& 67.302&720.56	\\
     	\texttt{wdb} 	&0.3385	&		\textbf{0.0043}&0.0344 &	0.0536&	0.0713	& 0.2999&	1.0912	\\

            \bottomrule
\end{tabular}
}
\label{Table-LU-11}
\end{center}
\end{table*}

 \begin{example}[Real data with outliers]\label{ex:real-data-outlier}\end{example}

  To observe the different influences of outliers on the real data  as for seven classifiers, we repeat the process of
 Example \ref{ex:real-data-no-outlier} on the adjusted data sets. A certain proportion of samples from original 10 UCI data sets are picked  and further   flipped their labels according to the label rate in each data set. In addition, considering the big time in training nonlinear SVMs, we select
 the $r=5, 10$ proportion in all comparative experiments.
  Average results of  the seven  classifiers are recorded in Table \ref{Table-LU-22} and Table \ref{Table-LU-33} respectively. Similarly, all 7 classifiers show the roughly equivalent performance on \mACC. Specifically, the 4 non-convex classifiers are a little slightly better than the other 3 convex rivals. Yet $L'_{0/1}$-ADMM consistently displays the property of strong sparsity of 0-1 loss, having the least SVs except ${\ramp}$ on $\texttt{bre}$ $(r=5)$ as the noises are added. Especially, $L'_{0/1}$-ADMM shows more  robustness on sparsity for it has the lowest increase of SVs  when ratio of the noise grows larger in all nonlinear classifiers.
\begin{remark}\end{remark}

The sparsity is a big issue to valuate the SVM classifier. The quantity of SVs
not only determines the memory space of
the learned predictor, but also the computation
cost of using it. A vast body of literatures have devoted to the task to improve the sparse performance of  SVM, such as [43-47]. Fortunately, the binary classifier SVM with 0-1 loss has the innate ability to scale down the potential SVs because for the geometrical and numerical evidence shown in this paper. May another reasonable mathematical explanation is that the binary property of 0-1 loss function, namely, consistent  punishment or assessment on all misclassified data points, brings about the overlapping of their representative function when choosing a few of them (SVs)  to express or construct decisive surface. Therefore, we can gradually reduce much redundant SV candidates until  a most economical SV set, just like maximal linear independent group of all SVs in [48], is sifted out.

\begin{table*}[htbp]
\begin{center}
\renewcommand\tabcolsep{2.5pt}
\caption{Comparisons of seven classifiers on 10 UCI data sets(r=5).}
\resizebox{4.5in}{!}
{
\begin{tabular}{lccccclccccc}
\toprule
&\multicolumn{6}{c}{\mACC~(\%)}\\
\text{Name} & $L'_{0/1}$ &{\libsvc}  &{\lssvc}  &{\l2svc} &{\ramp}&{\rsvc}&{\rshsvc} \\\hline
		\texttt{bre}  	& 92.45    &92.33	&92.58	&92.67	&92.50		&92.69	 &\textbf{92.83} \\
       \texttt{ech}  	&	89.05	&	88.16	&	87.70	&	88.46	&88.36&88.63 &\textbf{89.79}	\\
	    \texttt{hea}  	&	83.04	&	83.26	&	81.22	&	83.30	&	83.22& \textbf{83.52} &	82.96	\\
     	\texttt{hou}	&	 93.52 	&93.63 	&	\textbf{93.77} &	92.42 		&	 89.37& 93.56& 92.83	\\
		\texttt{hyp}	&	\textbf{93.34}	&	91.83	&	93.17	&	92.39	&	93.23& 92.30&92.15	\\
	    \texttt{mon}	&	89.70	&	89.05	&	86.09	&	88.29	&	86.14&\textbf{90.09}&89.06\\
        \texttt{pim}	&74.26	&	74.96	&	74.78	&	74.82	&	73.09&74.73 &\textbf{75.11}		\\		
     	\texttt{two}	&	\textbf{93.17}&	93.10	&	93.14	&	93.06	&	92.99&93.14 & 93.13		\\
		\texttt{wav}  	&85.57	&	86.05	&	\textbf{86.23}&	86.08	&	81.65& 85.72&	85.98	\\
     	\texttt{wdb} 	&93.79	&	94.25	& 	93.95  &	\textbf{94.50}&	93.73	& \textbf{94.50}&	94.45	\\\hline
&\multicolumn{5}{c}{\mNSV} \\
			 \text{Name} &$L'_{0/1}$ &{\libsvc}  &{\lssvc}  &{\l2svc} &{\ramp}&{\rsvc}&{\rshsvc} \\\hline
		\texttt{bre}  	& 127.56&132.86 	&	629.10	&	604.29	&\textbf{38.25}		&	130.53& 144.01 \\
       \texttt{ech}  	&	\textbf{8.24}	&	54.17	&	117.90	&	76.24	&	42.58&47.18 &	38.87	\\
	    \texttt{hea}  	&	\textbf{29.40}	&	146.04	&	243.00	&	226.54	&	217.14& 180.00 &	122.61	\\
     	\texttt{hou}	&	 \textbf{60.40}&107.93	&	391.50 	&	241.54		&	94.33& 186.93& 133.34\\
		\texttt{hyp}	&	\textbf{31.64}	&	548.70	&	2846.70	&	2243.36	&	339.26& 508.88&1068.58	\\
	    \texttt{mon}	&	\textbf{69.70}	&148.23	&388.80	&	225.75	&	280.92&199.83&156.57\\
        \texttt{pim}	&\textbf{206.57}	&	457.70	&	691.20	&	639.36	&343.11&400.94 &534.00		\\		
     	\texttt{two}	&\textbf{15.88}&	1897.10	&	6660.00	&	6660.00	&	1122.50&2385.70&1516.50		\\
		\texttt{wav}  	&\textbf{84.82}&	1706.80	&	4500.00	&	3538.20	&	1310.00& 1714.40&1331.80	\\
     	\texttt{wdb} 	&	\textbf{10.01}&	135.60	& 512.10	&	349.01&	107.87	& 160.38&	141.78	\\\hline
     &\multicolumn{5}{c}{\mCPU~(seconds)} \\
	   \text{Name} &$L'_{0/1}$     &{\libsvc}  &{\lssvc}   &{\l2svc}  &{\ramp} &{\rsvc} &{\rshsvc} \\\hline
		\texttt{bre}  	& 0.7095&\textbf{0.0142} 	&	0.0246	&	0.0558	&	0.2013	&	0.5612& 1.2262 \\
       \texttt{ech}  	&	0.0436&\textbf{0.0007}		&	0.0119	&	0.0056	&	0.0056&0.0201 &	0.3131	\\
	    \texttt{hea}  	&	0.1442&\textbf{0.0036}		&	0.0146	&	0.0099	&	0.0891& 0.0592 &	0.7212	\\
     	\texttt{hou}	&	0.1544&\textbf{0.0071}		&	0.0137	&	0.0148	&	0.0451& 0.1378 &	1.2288	\\
		\texttt{hyp}	&27.584	&\textbf{0.3208}		&	0.3901	&	3.1083	&45.270& 41.010&96.986	\\
	    \texttt{mon}	&	0.2561	&0.0232	&	\textbf{0.0161}	&	0.0269	&	0.0460&0.2706&2.0151\\
        \texttt{pim}	&1.3541	&	0.0277	&\textbf{0.0259}		&	0.0800	&0.601&0.6018 &1.5574		\\		
     	\texttt{two}	&122.62	&	\textbf{1.1296}	&	2.9587	&	19.777	&	65.807&184.55 & 431.13		\\
		\texttt{wav}  	&59.012	&	\textbf{0.8538}	&	1.1906	&	8.5727	&61.640& 54.856&912.79	\\
     	\texttt{wdb} 	&0.3517	&		\textbf{0.0086}&0.0700 &	0.0537&	0.1074	& 0.3235&	1.0270	\\
            \bottomrule
\end{tabular}
}
\label{Table-LU-22}
\end{center}
\end{table*}

\begin{table*}[htbp]
\begin{center}
\renewcommand\tabcolsep{2.5pt}
\caption{Comparisons of seven classifiers on 10 UCI data sets(r=10).}
\resizebox{4.5in}{!}
{
\begin{tabular}{lccccclccccc}
\toprule
&\multicolumn{6}{c}{\mACC~(\%)}\\
        \text{Name} & $L'_{0/1}$ &{\libsvc}  &{\lssvc}  &{\l2svc} &{\ramp}&{\rsvc}&{\rshsvc} \\\hline
		\texttt{bre}  	&\textbf{89.52} &88.27	&88.57	&88.63	&88.98		&88.42	 &88.82 \\
       \texttt{ech}  	&	87.30	&	87.19	&	85.75	&	87.79	&86.34& \textbf{88.65}&88.22	\\
	    \texttt{hea}  	&	\textbf{78.19}	&	77.81	&	76.37	&	78.00	&	77.00& 78.11 &	78.15	\\
     	\texttt{hou}	&	 89.26 	&89.24 	&	89.29 &89.25 		&\textbf{89.37}	&89.01& 88.82	\\
		\texttt{hyp}	&	88.21	&	87.49	&	88.41	&	87.46&\textbf{88.47}&87.60&87.12	\\
	    \texttt{mon}	&	81.98	&	82.58	&	76.70	&	76.70	&	75.77&82.04&\textbf{83.37}\\
        \texttt{pim}	&71.20	&	\textbf{71.93}	&	71.52	&	71.34	&	71.19&70.68 &71.39		\\		
     	\texttt{two}	&	88.19&	88.19	&	86.56	&	88.14	&	88.09&87.72 &\textbf{88.22} 	\\
		\texttt{wav}  	&81.63	&	81.91	&	81.71&	81.76	&	81.44& 81.40&\textbf{82.16}		\\
     	\texttt{wdb} 	&88.68	&	88.82	& 	88.28  &	88.43&	88.17	& 87.11&\textbf{88.84}		\\\hline
&\multicolumn{5}{c}{\mNSV} \\
			 \text{Name} &$L'_{0/1}$ &{\libsvc}  &{\lssvc}  &{\l2svc} &{\ramp}&{\rsvc}&{\rshsvc} \\\hline
		\texttt{bre}  	& \textbf{154.84}&199.89 	&	629.10	&	629.10	&311.67	&	257.69& 433.24 \\
       \texttt{ech}  	&	\textbf{10.16}	&	74.16	&	117.90	&	96.46	&	61.30&62.63 &	50.70	\\
	    \texttt{hea}  	&	\textbf{12.50}	&	148.66	&	243.00	&	77.00	&	78.11& 78.15 &	122.40	\\
     	\texttt{hou}	&	 \textbf{57.33}&156.92	&	391.50 	&	364.77		&	70.98&203.28& 141.00\\
		\texttt{hyp}	&	\textbf{35.90}	&	1255.43	&	2846.70	&	2596.97	&	424.53& 843.49&243.00	\\
	    \texttt{mon}	&	\textbf{23.60}	&230.49	&388.80	&	310.10	&	191.88&177.81&212.41\\
        \texttt{pim}	&\textbf{237.35}	&	480.50	&	691.20	&	686.20	&469.38&530.69 &491.32		\\		
     	\texttt{two}	&\textbf{13.98}&	3633.10	&	6660.00	&	6660.00	&	759.70&2592.51&2274.53		\\
		\texttt{wav}  	&\textbf{78.42}&	2107.27	&	4500.00	&	4229.15	&	942.31& 2106.00&2432.97	\\
     	\texttt{wdb} 	&	\textbf{33.99}&	210.16	& 512.10	&	465.39&	116.69	& 180.43&	227.31	\\\hline
&\multicolumn{5}{c}{\mCPU~(seconds)} \\
	   \text{Name} &$L'_{0/1}$     &{\libsvc}  &{\lssvc}   &{\l2svc}  &{\ramp} &{\rsvc} &{\rshsvc} \\\hline
		\texttt{bre}  	& 0.7295&\textbf{0.0185} 	&	0.0241	&	0.0430	&	0.2260	&	0.4232& 1.1956 \\
       \texttt{ech}  	&	0.0844&\textbf{0.0008}		&	0.0121	&	0.0033	&	0.0060&0.0323 &	0.1975	\\
	    \texttt{hea}  	&	0.1056&\textbf{0.0033}		&	0.0147	&	0.0077	&	0.0301& 0.0750 &	0.5020	\\
     	\texttt{hou}	&	0.1703&\textbf{0.0100}		&	0.0212	&	0.0204	&	0.0419& 0.2893 &	0.2323	\\
		\texttt{hyp}	&29.580	&\textbf{0.3944}		&	0.3893	&	2.5705	&70.899& 34.896&153.752	\\
	    \texttt{mon}	&	0.2610	&	\textbf{0.0121}&0.0139	&	0.0209	&	0.0660&0.2185&0.9130\\
        \texttt{pim}	&1.3720	&	0.0271	&\textbf{0.0262}		&	0.0597	&0.2932&0.5257 &2.7212		\\		
     	\texttt{two}	&130.76	&	\textbf{2.0902}	&	2.8031	&	19.696	&	69.187&302.50 & 475.86		\\
		\texttt{wav}  	&56.837	&1.9679	&\textbf{1.1997}	&	8.6840	&47.324& 60.727&451.11	\\
     	\texttt{wdb} 	&0.1004	&		\textbf{0.0113}&0.0359 &	0.0516&	0.1173&0.4707&	1.1446	\\
            \bottomrule
\end{tabular}
}
\label{Table-LU-33}
\end{center}
\end{table*}
\section{Conclusion}\label{fnlsvm0888}

In this paper, we have explored the  $L_{0/1}$-KSVM for the nonlinear SVM with 0-1 loss soft margin, the build-up of framework on its optimal condition and algorithm $L'_{0/1}$-ADMM followed the success of $L_{0/1}$-SVM, an efficient algorithm on its linear formulation. In the proposed $L_{0/1}$-KSVM, the SVs, a key index to evaluate the performance of SVM, showed nice geometric and numerical properties. All SVs just fell into the parallel decision surfaces, further, the experiment manifested that the number of SVs occupied the enormous superiority in  compared models. Obviously, the property of having few SVs greatly  improves the prediction speed because  the number of SVs basically determines the overall efficiency of the SVM as a classifier. However, suffering from the curse of kernelization, the long computation  time exhausted in training $L_{0/1}$-KSVM is a flaw for its algorithm and finding an effective kernel trick to scale up the proposed algorithm is our task for the future.

 \section*{Acknowledgements}
This work is supported by the Natural Science Foundation of Hainan Province (Nos.120RC449,  620QN234), the National Natural Science Foundation of China (Nos.12271131, 61866010, 11871183, 62066012),  the Scientific Research cultivation project for young teachers at Hainan University (No.RZ2100004477),   the Education Department of Hainan Province\\ (No.Hnky2020-3), the Key Laboratory of Engineering Modeling and Statistical Computation of Hainan Province. Moreover, great appreciation needs sending to Doctor Hua-jun Wang and  Professor Nai-hua Xiu  for their valuable suggestions.

\section{References}

[1] C. Cortes and V. Vapnik, Support vector networks, Machine Learning. 20 (3) (1995)  273-297.

[2] C. J. Burges,  A tutorial on support vector machines for pattern recognition, Data mining and knowledge discovery.   2(2) (1998)   121-167.

[3] N. Deng, Y. Tian, and C. Zhang,  Support vector machines: optimization based theory, algorithms, and extensions,  CRC press, Taylor \& Francis Group, New York, 2012.

[4] B. K. Natarajan, Sparse approximate solutions to linear systems,	SIAM journal on computing 24(2) (1995) 227-234.

[5]  E. Amaldi and V. Kann, On the approximability of minimizing nonzero variables or unsatisfied relations in linear systems, Theoretical Computer Science. 209(1) (1998) 237-260.

[6]  S. Ben-David, N. Eiron, P. M. Long, On the Difficulty of Approximately Maximizing Agreements, Journal of Computer and System Sciences. 66(3) (2003) 496-514.

[7] V. Feldman, V. Guruswami, P. Raghavendra, et al, Agnostic Learning of Monomials by Halfspaces is Hard, in: Proceedings of the 50th IEEE Symposium on Foundations of Computer
Science, 2009.

[8] T. Zhang,  Statistical behavior and consistency of classification methods based on convex risk
minimization, The Annals of Statistics. 32(1) (2004) 56-85.

[9] J. A. K. Suykens and J. Vandewalle, Least squares support vector machine classifiers,  Neural processing letters. 9(3) (1999)  293-300.
[10] T. Zhang and F. J. Oles,  Text categorization based on regularized
linear classification methods,  Information Retrieval.   4(1) (2008)  5-31.

[11] V. Jumutc, X. Huang, and J. A. K. Suykens,  Fixed-size pegasos for hinge and pinball loss SVM, in: The 2013 International Joint Conference on Neural Networks (IJCNN). 2013, pp. 1-7.

[12] J. Friedman, T. Hastie, and R. Tibshirani,  Additive
logistic regression: a statistical view of boosting, The Annals of Statistics. 28(2) (2000)  337-374.

[13] R. Collobert, F. Sinz, J. Weston, et al,  Trading convexity for scalability, in: Proceedings of the 23rd international conference on Machine learning, 2006,  201-208.

[14]  H. Wang, Y. Shao and N. Xiu,  Proximal operator and optimality conditions for ramp loss SVM.  Optimization Letters 16(3)  (2022) 999-1014.

[15] G. Xu, Z. Cao, B. G. Hu, et al,  Robust support vector machines based on the rescaled hinge loss function, Pattern Recognition. 68(2017) 139-148.

[16] Y. L. Feng, Y. N. Yang, X. L. Huang, S. Mehrkanoon, and J. A. K. Suykens,  Robust support vector machines for classification with nonconvex and smooth losses, Neural computation. 28(6) (2016) 1217-1247.

[17] Perez-Cruz, Fernando, et al,  Empirical risk minimization for support vector classifiers, IEEE Transactions on Neural Networks. (2003)  296-303.

[18] Shalev-Shwartz, Shai, Ohad Shamir, and Karthik Sridharan, Learning linear and kernel predictors with the 0-1 loss function, in: Twenty-Second International Joint Conference on Artificial Intelligence, 2011.

[19] Orsenigo, Carlotta, and Carlo Vercellis, Multivariate classification trees based on minimum features discrete support vector machines, IMA Journal of Management Mathematics. 14(2) (2003) 221-234.

[20] H. Wang, Y. Shao, S. Zhou, C. Zhang, and N. Xiu, Support Vector Machine Classifier via $ L\_ $\{$0/1$\}$ $ Soft-Margin Loss, IEEE Transactions on Pattern Analysis and Machine Intelligence. (2021).  DOI: 10.1109/TPAMI.2021.3092177.

[21] Y. Shao, et al, Key issues of support vector machines and future prospects, Scientia Sinica Mathematica. 50(9) (2020) 1233.

[22] A. Rahimi, B. Recht,  Random features for large scale kernel machines, Advances in Neural Information Processing Systems 20, in: Proceedings of the Twenty-First Annual Conference on Neural Information Processing Systems, 2007.

[23] F. Liu, X. Huang, Y. Chen, et al, Random Features for Kernel Approximation: A Survey in Algorithms, Theory, and Beyond,  IEEE Transactions on Pattern Analysis and Machine Intelligence. 44(10) (2021) 7128-7148.

[24] J. Mercer,  Functions of positive and negative type, and their connection with the theory of integral equations, Proceedings of the Royal Society A Mathematical Physical and  Engineering Sciences. 83(559) (1909) 69-70.

[25] T. Poggio,  On optimal nonlinear associative recall, Biological Cybernetics.  19(4) (1975) 201-209.

[26] T. S. Jaakkola, D. Haussler,  Probabilistic kernel regression models, in: Proceedings of the 1999
Conference on AI and Statistics, 1999.

[27] T. Hofmann, B. Scholkopf and A. J. Smola,  Kernel methods in machine learning, Annals of Statistics. 36(3) (2008) 1171-1220.

[28] G. Cybenko, Approximation by superpositions of a sigmoidal function, Mathematics of Control, Signals and Systems. 2(4) (1989) 303-314.

[29] N. Aronszajn,  Theory of reproducing kernels,  Transactions of the American Mathematical Society. 68(3) (1950) 337-404.

[30] Rudin, Walter, Real and complex analysis,  Tata McGraw-hill education, New York, 2006.

[31] P. G. Wahba,  Spline models for observational data, Journal of Approximation Theory. 66(3) (1991) 354-354.

[32] B. Scholkopf, R. Herbrich, A. J. Smola,  A generalized representer theorem, in: International conference on computational learning theory, 2001,  416-426.

[33] M. R. Hestenes, E. L. Stiefel,  Methods of Conjugate Gradients for Solving Linear Systems, Journal of Research of the National Bureau of Standards. 49(6) (1952) 409-436.

[34] G. B. Ye, Y. F. Chen, and X. H. Xie, Efficient variable selection in support vector machines via the alternating direction method of multipliers, in: Proceedings of the Fourteenth International Conference on Artificial Intelligence and Statistics, 2011, 832-840.

[35] L. Guan, L. B. Qiao, D. S. Li, T. Sun, K. S. Ge, and X. C. Lu,  An efficient ADMM-based algorithm to nonconvex penalized support vector machines, in: 2018 IEEE International Conference on Data Mining Workshops (ICDMW), 2018,  1209-1216.

[36] M. Hong, Z.Q. Luo, and M. Razaviyayn,  Convergence analysis of
alternating direction method of multipliers for a family of nonconvex
problems,  SIAM Journal on Optimization. 26(1) (2016) 337-364.

[37] B. Jiang, T. Lin, S. Ma, et al,  Structured nonconvex and nonsmooth optimization: algorithms and iteration complexity analysis, Computational Optimization and Applications. 72(1) (2019) 115-157.

[38] Z. Jia, J. Huang,  Wu,  An incremental aggregated proximal ADMM for linearly constrained nonconvex optimization with application to sparse logistic regression problems, Journal of Computational and Applied Mathematics.  390 (2021),  113384.

[39] J. Nocedal, S. Wright,  Numerical Optimization, NewYork, Springer, 2006.

[40] Z. Wang, K. Crammer, S. Vucetic, Multi-Class Pegasos on a Budget, in: International Conference on International Conference on Machine Learning, 2010.

[41]  C. C. Chang, C. J. Lin,  LIBSVM: a library for support vector machines, ACM transactions on intelligent systems and technology (TIST). 2(3) (2011) 27.

[42] K. Pelckmans, J. A. K. Suykens, T. V. Gestel, J. D. Brabanter, L. Lukas, B. Hamers, B. D. Moor, and J. Vandewalle,  LSSVM lab: a matlab/c toolbox for least squares support vector machines,  Tutorial. KULeuven-ESAT. Leuven, Belgium, 142(2002)  1-2.

[43] M.  Wu, B. Scholkopf, and G. Bakir,  Building sparse
large margin classifiers, in: Proceedings of the Twenty-Second International Conference, 2005,  996-1003.

[44] K. Huang, R. Zheng, R. Sun, R. Hotta, R. Fujimoto,  R. Naoi,  Sparse learning for support vector classification,  Pattern Recognition Letters. 31(13) (2010) 1944-1951.

[45] A. Cotter, S. Shalev-Shwartz, N. Srebro,  Learning optimally sparse support vector machines, in: 30th International
Conference on Machine Learning, 2013, 266-274.

[46] M. Aliquintuy, E. Frandi, R. Nanculef, J.A.K. Suykens, Efficient Sparse Approximation of Support Vector Machines Solving a Kernel Lasso,
Lecture Notes in Computer Science.  10125 (2017)   208-216.

[47] Z. Liu, D. Elashoff, S. Piantadosi,  Sparse support vector machines with $l_0$ approximation for ultra-high dimensional omics data,  Artificial Intelligence in Medicine. 96   (2019)  134-141.

[48] T. Downs, K. E. Gates, and A. Masters,  Exact simplification of support
vector solutions,  Journal of Machine Learning Research. 2(2001)  293-297.

[49] B. Mordukhovich and N. Nam, An easy path to convex analysis and applications,  Morgan and Claypool Pubulishers, Kentfield,  2014.

%\vspace*{-0.01in}
%%\vspace*{-0.3in}
%\noindent
%\rule{12.6cm}{.1mm}

%\section*{Biographical Sketch and Photo}
%
%Upon acceptance of an article, a brief biographical sketch and
%photograph of each author are to be supplied to the Publisher.
%
%\biophoto{wang}{{\bf Chuan-Cheng Wang} received the
%B.S.~degree\break
%in electrical engineering from National Sun Yat-Sen University,
%Kaoh-\break siung, Taiwan in 1992 and the M.S. degree in
%computer science from National Chiao Tung University, Hsinchu,\break
%Taiwan in 2001.}
%
%\vglue-1.75truein
%\hspace*{2.45truein}
%\biophoto{gao}{{\bf Yongsheng Gao} received the B.Sc. and\break
%M.Sc. degrees in electronic engineering from\break Zhejiang
%University,\break China, in 1985 and 1988 respectively, and the
%Ph.D. in computer engineering from Nanyang Technological University,
%Singapore. Currently, he is an assistant professor with Nanyang
%Technological University, Singapore. }
\appendices
\section{Proofs of all theorems}

\subsection{Proof of Proposition \ref{Dtheorem}}\label{proof222}
\textbf{Proof.}
Now suppose the kernel function has the trivial form, namely, $k(\widetilde{\textbf{x}_i},\widetilde{\textbf{x}_j})=\left \langle \widetilde{\textbf{x}_i},\widetilde{\textbf{x}_j} \right \rangle({ i, j\in{\mathbb{N}}_{m}})$ in new model \eqref{gksvmnew}, then ${\alpha}^T \widetilde{\bf{K}}{\alpha}=\frac{1}{2} \Vert{\textbf{w}} \Vert^2+\frac{1}{2} b^2$ by denoting $-\sum_{i=1}^m {\alpha_i}{y_i}{{\widetilde{\textbf{x}}_{i}}}=({\textbf{w}}^{\top}, b)^{\top}$. Consequently,  nonlinear $L_{0/1}$-KSVM \eqref{gksvmnew} has successfully degenerated into  linear $L_{0/1}$-SVM \eqref{reformulate} with an extra regularization $\frac{1}{2} b^2$ in its objective function. \hfill   $ \Box $

\subsection{Proof of Theorem \ref{gol-p}}\label{proof22}

\subsubsection{A lemma for proof of Theorem \ref{gol-prox2}~(i)}

\begin{lemma} \label{globalnew}
Suppose $g({\textbf{u}})$ is a gradient continuous Lipschitz function with Lipschitz constant $L$ and ${\textbf{u}}^*$ is a global minimizer to
\begin{eqnarray} \label{inverse}
\min_{{\textbf{u}}\in {\mathbb{R}}^{m}} ~~ g({\textbf{u}})+C\|{\textbf{u}}_{+}\|_{0},
\end{eqnarray}
 then
\begin{eqnarray}\label{equation}
{\textbf{u}}^*&=\text{Prox}_{{\gamma}C\|({\cdot})_{+}\|_{0}}({\textbf{u}}^*-{\gamma}\nabla{g({\textbf{u}}^*)}
\end{eqnarray}
for  ${\gamma} \in (0,1/L)$.
\end{lemma}

\textbf{Proof.} Denote $\textbf{z}:=\text{Prox}_{{\gamma}C\|({\cdot})_{+}\|_{0}}({\textbf{u}}^*-{\gamma}\boldsymbol{\lambda^*})$ with $\boldsymbol{\lambda^*}:=\nabla{g({\textbf{u}}^*)}$. We will prove $\textbf{u}^*=\textbf{z}$.
{\allowdisplaybreaks
\begin{eqnarray}\label{relation}\nonumber
&&g({\textbf{u}}^*)+C\|{\textbf{u}}^*_{+}\|_{0}
\nonumber\\
&{\leq}&g({\textbf{z}})+C\|{\textbf{z}}_{+}\|_{0}\nonumber\\
&{\leq}& g({\textbf{u}}^*)+ \langle\boldsymbol{\lambda^*},{\textbf{z}}-{\textbf{u}^*}\rangle+\frac{L}{2}\| {\textbf{z}}-{\textbf{u}^*} \|^2+C\|{\textbf{z}}_{+}\|_{0}\nonumber\\
&{\leq}& g({\textbf{u}}^*)+ \langle\boldsymbol{\lambda^*},{\textbf{z}}-{\textbf{u}^*}\rangle+\frac{L}{2}\| {\textbf{z}}-{\textbf{u}^*} \|^2+C\|{\textbf{u}}^*_{+}\|_{0}\nonumber\\
&+&\frac{1}{2\gamma}\| {\textbf{u}}^*-({\textbf{u}^*}-\gamma\boldsymbol{\lambda^*}) \|^2-\frac{1}{2\gamma}\| {\textbf{z}}-({\textbf{u}^*}-\gamma\boldsymbol{\lambda^*}) \|^2\nonumber\\
&{\leq}& g({\textbf{u}}^*)+C\|{\textbf{u}}^*_{+}\|_{0} +\frac{L-1/\gamma}{2} \| {\textbf{z}}-{\textbf{u}^*} \|^2\nonumber,\nonumber
\end{eqnarray}}
which indicates $\| {\textbf{z}}-{\textbf{u}^*} \|^2\leq 0$ due to $0<{\gamma}<1/L$, hence $\textbf{u}^*=\textbf{z}$. \hfill   $ \Box $

\subsubsection{Proof of Theorem \ref{gol-prox2}~(i)}

Firstly, since we assume that $\widetilde{\bf{K}}$ is invertible, the formulation \eqref{gksvmnew} of { $L_{0/1}$-KSVM} can be transformed as minimizer problem only containing single variable $\textbf{u}$:
\begin{eqnarray} \label{inverse2}
\min_{{\textbf{u}}\in {\mathbb{R}}^{m}} ~~ ({\textbf{u}}-{\textbf{e}})^{\top}{\widetilde{\bf{K}}^{-1}}({\textbf{u}}-{\textbf{e}})+C\|{\textbf{u}}_{+}\|_{0},
\end{eqnarray}
In Lemma \ref{globalnew}, denoting $L=\lambda_{\max}(\widetilde{\bf{K}}^{-1})$ and $g({\textbf{u}}):=({\textbf{u}}-{\textbf{e}})^{\top}{\widetilde{\bf{K}}^{-1}}({\textbf{u}}-{\textbf{e}})$,  we obtain $\nabla{g(\textbf{u})}={\widetilde{\bf{K}}^{-1}}({\textbf{u}}-{\textbf{e}})$. Moreover, $\textbf{u}^*$ is the global minimizer of \eqref{inverse2} by considering that $( \textbf{u}^*;\alpha^*)$ is the global minimizer of \eqref{gksvmnew}.
Now the Lemma \ref{globalnew} can lead us the conclusion as follows:
\begin{eqnarray}\label{gradient10}
 {\textbf{u}}^*&=\text{Prox}_{\gamma C\|(\cdot)_{+}\|_{0}}({\textbf{u}}^*-\gamma {\widetilde{\bf{K}}^{-1}}({\textbf{u}^*}-{\textbf{e}})),
\end{eqnarray}
for any $0<\gamma<1/\lambda_{\max}(\widetilde{\bf{K}}^{-1})$. Next by letting ${{\alpha}^*}= {\widetilde{\bf{K}}^{-1}}({\textbf{u}^*}-{\textbf{e}})$, then
\begin{eqnarray}\label{gradient20}
\widetilde{\bf{K}}{{\alpha}^*}=\widetilde{\bf{K}}{\widetilde{\bf{K}}^{-1}}({\textbf{u}^*}-{\textbf{e}})={\textbf{u}^*}-{\textbf{e}},
\end{eqnarray}
Hence, combination of \eqref{gradient10} and \eqref{gradient20} completes the proof. \hfill   $ \Box $

\subsubsection{Proof of Theorem \ref{gol-prox2} (ii)}

Suppose $\boldsymbol{\phi}^*:= ( \alpha^*; \textbf{u}^*)$ is a P-stationary point of \eqref{gksvmnew} with $\gamma>0$, then we have
\begin{equation}
\label{aaalfi-0}\begin{cases}&\textbf{u}^*-{\widetilde{\textbf{K}}} \alpha^*={\textbf{e}},\\
                            &\text{Prox}_{\gamma C\|({\cdot})_{+}\|_{0}}(\textbf{u}^*-\gamma{\alpha}^*)=\textbf{u}^*,
\end{cases}.
\end{equation}

 Let $\Theta$ be the feasible region of \eqref{gksvmnew}, namely,
 \begin{eqnarray}\label{feasible}
 \Theta:=\{\boldsymbol{\phi}:=( \alpha; \textbf{u}): \textbf{u} -\widetilde{\bf{K}}  \alpha=
{\textbf{e}}\}.
 \end{eqnarray}
 Furthermore, the function $||\textbf{u}_+||_0$  is lower semi-continuous at $\boldsymbol{\phi}^*\in \Theta $, then by  [49, Proposition 4.3], there is a neighborhood $U(\boldsymbol{\phi}^*,\delta_1)$ of $\boldsymbol{\phi}^*\in \Theta $ with $\delta_1>0$ such that
\begin{eqnarray}\label{Uplas}
||{\textbf{u}}_+||_0> ||{\textbf{u}}^*_+||_0-\frac{1}{2},\forall \boldsymbol{\phi}\in \Theta\cap U(\boldsymbol{\phi}^*,\delta_1).
\end{eqnarray}
In addition, since $ {\alpha}^T \widetilde{\bf{K}}{\alpha} $ is locally Lipschitz continuous in ${\mathbb{R}}^{m}$, there exists a neighborhood $U(\boldsymbol{\phi}^*,\delta_2)$ of $\boldsymbol{\phi}^*\in \Theta$ with $\delta_2>0$ such that
  \begin{eqnarray}\label{such}
| {\alpha}^T \widetilde{\bf{K}}{\alpha}  - {{\alpha}^*}^T \widetilde{\bf{K}}{{\alpha}^*}  |\leq 2C,~~ \forall \boldsymbol{\phi}\in \Theta\cap U(\boldsymbol{\phi}^*,\delta_2).
  \end{eqnarray}

Denote $\delta=\min\{\delta_1,\delta_2\}$. Now we show that $\boldsymbol{\phi}^*$ is a local minimizer of \eqref{gksvmnew}. Namely, there exists a neighborhood $U(\boldsymbol{\phi}^*,\delta)$ of $\boldsymbol{\phi}^*\in \Theta$ with $\delta>0$ such that
\begin{eqnarray}\label{local}
\frac{1}{2} {{\alpha}^*}^T \widetilde{\bf{K}}{{\alpha}^*}+C\|{\textbf{u}}^*_{+}\|_{0}&\leq&\frac{1}{2} {\alpha}^T \widetilde{\bf{K}}{\alpha}+C\|{\textbf{u}}_{+}\|_{0},\\ \nonumber
 &&\forall \boldsymbol{\phi}\in \Theta\cap U(\boldsymbol{\phi}^*,\delta).
\end{eqnarray}
 For this purpose, let $\Gamma_*:=\{i: u_i^* = 0\}$ and $\overline{\Gamma}_*:=\mathbb{N}_m \backslash \Gamma_*.$ It follows from  \eqref{proxm1} that we obtain
 %\begin{eqnarray}\label{lamma}
%\bm{\lambda}^*_{\overline{\Gamma}_*}=0.
% \end{eqnarray}
%the P-stationary point $\bm{\phi}^*$  satisfies \eqref{prox-G}. So we have
 \begin{eqnarray}\label{fact-lambda}
\begin{cases}
-\sqrt{2C/\gamma}\leq\alpha_i^*\leq 0,~u_i^*=0,~~&\forall i \in \Gamma_*,\\
\alpha_i^*=0,~u_i^*\neq0,~~&\forall i \in \overline {\Gamma}_*.
\end{cases}
\end{eqnarray}

Based on these, we consider a local region  $\Theta_1$ of $\Theta$ as
 \begin{eqnarray}\label{loc-feasible}
 \Theta_1:=\Theta\cap\{\boldsymbol{\phi}: u_i\leq0,~~ i\in \Gamma_* \}.
 \end{eqnarray}
We split the proof of the (\ref{local}) into
the following two cases:

{\bf Case (i):} $\boldsymbol{\phi}\in\Theta_1\subseteq \Theta$ and $\boldsymbol{\phi}\in U(\boldsymbol{\phi}^*,\delta).$  It is easy to see that $\boldsymbol{\phi}^*\in \Theta_1$ by \eqref{aaalfi-0}. Then for any $\boldsymbol{\phi}\in \Theta_1$, we have   two facts
 \begin{eqnarray}\label{fact-u-T}u_i\leq0,~i\in \Gamma_*  \end{eqnarray}

 and $\textbf{u} -\widetilde{\bf{K}}  \alpha=
{\textbf{e}}$, which and \eqref{aaalfi-0} suffice  to
 \begin{eqnarray}\label{fact-u-u*}
\widetilde{\bf{K}}( \alpha- \alpha^*) = \textbf{u}-\textbf{u}^* .
 \end{eqnarray}

The following chain of inequalities hold for any $\boldsymbol{\phi}\in \Theta_1$,
\begin{eqnarray}
&&\frac{1}{2} {\alpha}^T \widetilde{\bf{K}}{\alpha}  -\frac{1}{2} {{\alpha}^*}^T \widetilde{\bf{K}}{{\alpha}^*}\nonumber\\
&\geq&\frac{1}{2} {\alpha}^T \widetilde{\bf{K}}{\alpha} -\frac{1}{2} ({\alpha}-{\alpha}^*)^T \widetilde{\bf{K}}({\alpha}-{\alpha}^*) -\frac{1}{2} {{\alpha}^*}^T \widetilde{\bf{K}}{{\alpha}^*}\nonumber\\
&=&{{\alpha}^*}^T\widetilde{\bf{K}}{\alpha}  - {{\alpha}^*}^T \widetilde{\bf{K}}{{\alpha}^*}\nonumber\\
&\overset{\eqref{fact-u-u*}}{=}&\langle {\alpha}^*,{\textbf{u}-\textbf{u}^* } \rangle\nonumber\\
&=&\langle  {\alpha}^*_{\Gamma_* },\textbf{u}_{\Gamma_*}-\textbf{u}^*_{\Gamma_* }\rangle+\langle {\alpha}^*_{\overline {\Gamma}_* }, \textbf{u}_{\overline \Gamma_* }-\textbf{u}^*_{\overline \Gamma_*}\rangle\nonumber\\
&\overset{\eqref{fact-lambda}}{=}&\langle {\alpha}^*_{\Gamma_* },\textbf{u}_{\Gamma_*} \rangle\nonumber\\
\label{is}& \overset{\eqref{fact-lambda},\eqref{fact-u-T}}{\geq}&   0.
\end{eqnarray}
Since $||\textbf{u}_+||_0$ can only take  values from $\{0,1,\cdots,m\}$, this together with \eqref{Uplas} allows us to conclude that
\begin{eqnarray}\label{inequality}
||{\textbf{u}}_+||_0\geq ||{\textbf{u}}^*_+||_0,\forall \boldsymbol{\phi}\in \Theta\cap U(\boldsymbol{\phi}^*,\delta_1).
\end{eqnarray}
Therefore, for any $\boldsymbol{\phi}\in \Theta_1\cap U(\boldsymbol{\phi}^*,\delta)\subseteq \Theta_1\cap U(\boldsymbol{\phi}^*,\delta_1)$, then  \eqref{is} and \eqref{inequality} lead to
\begin{eqnarray}\label{have}
\frac{1}{2} {{\alpha}^*}^T \widetilde{\bf{K}}{{\alpha}^*}+C\|{\textbf{u}}^*_{+}\|_{0}&\leq&\frac{1}{2} {\alpha}^T \widetilde{\bf{K}}{\alpha}+C\|{\textbf{u}}_{+}\|_{0}.\\ \nonumber.
\end{eqnarray}
{\bf Case (ii):} $\boldsymbol{\phi} \in (\Theta\setminus\Theta_1)$ and $\boldsymbol{\phi}\in  U(\boldsymbol{\phi}^*,\delta)$.  For any $\boldsymbol{\phi} \in (\Theta\setminus\Theta_1)$, we claim that there exists $i_0\in \Gamma_* $ with $u_{i_0}^*=0$ but $u_{i_0}>0$, which implies $\|(u_{i_0}^*)_{+}\|_{0}=0$ but
$\|(u_{i_0})_{+}\|_{0}=1$. By $\boldsymbol{\phi}\in  U(\boldsymbol{\phi}^*,\delta)$ and \eqref{inequality}, we have
\begin{eqnarray}\label{we}
\|{\textbf{u}}_{+}\|_{0}\geq \|{\textbf{u}}^*_{+}\|_{0}+1.
\end{eqnarray}

This together with \eqref{such} obtains that for  any $\boldsymbol{\phi}\in (\Theta\setminus\Theta_1)\cap U(\boldsymbol{\phi}^*,\delta),$
 \begin{eqnarray} \label{any}
\frac{1}{2} {{\alpha}^*}^T \widetilde{\bf{K}}{{\alpha}^*}+C\|{\textbf{u}}^*_{+}\|_{0}&\leq& \frac{1}{2} {{\alpha}^*}^T \widetilde{\bf{K}}{{\alpha}^*}+ C\|\textbf{u}_{+}\|_{0}-C \nonumber\\
 &\leq& \frac{1}{2} {{\alpha}}^T \widetilde{\bf{K}}{{\alpha}}+C\|\textbf{u}_{+}\|_{0}.
 \end{eqnarray}
Summarizing \eqref{have} and \eqref{any},  we obtain that $\boldsymbol{\phi}^*$ is a local minimizer of \eqref{gksvmnew} in a local region $\Theta \cap U(\boldsymbol{\phi}^*,\delta)$. The proof is completed.\hfill $\Box $

\subsection{Proof of Theorem \ref {convergence}}\label{conver1}
Since $T_k\subseteq \mathbb{N}_m$ has finite many elements, for sufficient large $k$,  there is a subset $J\subseteq\{1,2,3,\cdots\}$ such that
\begin{eqnarray}
\label{TKT}
T_j\equiv:T, ~~\forall~j\in J.
\end{eqnarray}
For notational simplicity, denote the sequence $\boldsymbol{\Psi}^k:=(\alpha^k,{\textbf{u}}^k,{\boldsymbol{\lambda}}^k)$ and its limit point $\boldsymbol{\Psi}^*:=( \alpha^*,{\textbf{u}}^*,{\boldsymbol{\lambda}}^*)$, namely $\{\boldsymbol{\Psi}^k\}\rightarrow \boldsymbol{\Psi}^*$. This also indicates $\{\boldsymbol{\Psi}^j\}_{j\in J}\rightarrow \boldsymbol{\Psi}^* $ and $\{\boldsymbol{\Psi}^{j+1}\}_{j\in J}\rightarrow \boldsymbol{\Psi}^*$. Taking the limit along with $J$ of \eqref{lambda-k}, namely, $k\in J, k\rightarrow\infty$, we have
\begin{eqnarray}\label{lambda-k-1}
\left\{
\begin{array}{lll}
{\boldsymbol{\lambda}}^*_{T}&=& {\boldsymbol{\lambda}}^*_{T}+\eta\sigma{\boldsymbol{\varpi}}^*_{T},\\
{\boldsymbol{\lambda}}^*_{\overline{T}}&=&\textbf{0},
\end{array}\right.
\end{eqnarray}
which derives ${\boldsymbol{\varpi}}^*_{T}= \textbf{0}$. Calculating the limit along with $J$ of \eqref{prox_u} and $\textbf{z}^k$ respectively yields
\begin{eqnarray}
\textbf{z}^*&=&{\textbf{e}}+\widetilde{\bf{K}} \alpha^*- {\boldsymbol{\lambda}}^*/\sigma\nonumber\\
&=&{\textbf{e}}+\widetilde{\bf{K}} \alpha^*-\textbf{u}^*+\textbf{u}^*- {\boldsymbol{\lambda}}^*/\sigma\nonumber\\
\label{zk-1}&=&- {\boldsymbol{\varpi}}^*+\textbf{u}^*- {\boldsymbol{\lambda}}^*/\sigma
\end{eqnarray}
and thus
\begin{eqnarray}
\label{prox11-0}{\textbf{u}}^*_{T}&=& {\bf0},\\
\label{prox11-1}{\textbf{u}}^*_{\overline{T}}&=& \textbf{z}^*_{\overline{T}}\\\nonumber
 &\overset{\eqref{zk-1}}{=}& - {\boldsymbol{\varpi}}^*_{\overline{T}}+\textbf{u}^*_{\overline{T}}- {\boldsymbol{\lambda}}^*_{\overline{T}}/\sigma\\
&\overset{\eqref{lambda-k-1}}{=}& - {\boldsymbol{\varpi}}^*_{\overline{T}}+\textbf{u}^*_{\overline{T}}.\nonumber
\end{eqnarray}
This proves  ${\boldsymbol{\varpi}}^*_{\overline{T}}=\textbf{0}$ and hence ${\boldsymbol{\varpi}}^*=\textbf{0}$. Again by \eqref{zk-1}, we obtain $\textbf{z}^*=\textbf{u}^*- {\boldsymbol{\lambda}}^*/\sigma$, which together with \eqref{prox11-0}, \eqref{prox11-1} and the definition of $L_{0/1}$ proximal operator indicates
\begin{eqnarray}\label{prox11-2}{\textbf{u}}^*=\text{Prox}_{\frac{C}{\sigma}\|(\cdot)_+\|_0}(\textbf{z}^*)=\text{Prox}_{\frac{C}{\sigma}\|(\cdot)_+\|_0}(\textbf{u}^*- {\boldsymbol{\lambda}}^*/\sigma).\end{eqnarray}

In addition, by taking the limit along with $J$ of \eqref{euqation-w}, we have
\begin{eqnarray*}
 ( \widetilde{\bf{K}} +\sigma \widetilde{\bf{K}}_{T}^\top \widetilde{\bf{K}}_{T}  ) \alpha^*&=& \sigma \widetilde{\bf{K}}_{T}^\top \textbf{v}_{T}^* \\
 &=& \sigma \widetilde{\bf{K}}_{T}^\top ({\textbf{u}}^*_{T}-{\textbf{e}}_{T}+{\boldsymbol{\lambda}}^*_{T}/\sigma)\\
 &=& \sigma \widetilde{\bf{K}}_{T}^\top ({\boldsymbol{\varpi}}^*_{T}+\widetilde{\bf{K}}_{T}\alpha^*+{\boldsymbol{\lambda}}^*_{T}/\sigma)\\
  &=& \sigma \widetilde{\bf{K}}_{T}^\top ( \widetilde{\bf{K}}_{T}\alpha^*+{\boldsymbol{\lambda}}^*_{T}/\sigma),
\end{eqnarray*}
where $\textbf{v}^*={\textbf{u}}^*-{\textbf{e}}+{\boldsymbol{\lambda}}^*/\sigma$ and the last two equations hold due to ${\boldsymbol{\varpi}}^*={\textbf{u}}^*-{\textbf{e}}-\widetilde{\bf{K}} \alpha^* =\textbf{0}$. The last equation suffices to that
\begin{eqnarray*}\label{euqation-w-1}
 \widetilde{\bf{K}} \alpha^*&=&\widetilde{\bf{K}}_{T}^\top  {\boldsymbol{\lambda}}^*_{T} \overset{\eqref{lambda-k-1}}{=} \widetilde{\bf{K}} ^\top  {\boldsymbol{\lambda}}^*=\widetilde{\bf{K}}{\boldsymbol{\lambda}}^*.
\end{eqnarray*}

Then a fabulous result is  $\alpha^*={\boldsymbol{\lambda}}^*$ for the symmetry and invertibility of $\widetilde{\bf{K}}$. Overall, we have
\begin{eqnarray*}
\left\{
\begin{array}{rll}
\textbf{u}^*-\widetilde{\bf{K}}\alpha^*&=&{\textbf{e}},\\
\text{Prox}_{\frac{C}{\sigma}\|(\cdot)_+\|_0}(\textbf{u}^*- {{\alpha}^*}/\sigma)&=&\textbf{u}^*.
\end{array}\right.
\end{eqnarray*}
Namely, $( \alpha^*; \textbf{u}^* )$ is a  P-stationary  point of problem \eqref{gksvmnew} where $\gamma=1/\sigma$. Then by Theorem \ref{gol-p} (ii), it is a local optimal solution of problem \eqref{gksvmnew}, which completes the proof.\hfill   $ \Box $
\end{document}